\documentclass{article} 
\usepackage{iclr2027_conference,times}

\usepackage{amsmath,amsfonts,bm}

\def\eqref#1{equation~\ref{#1}}

\def\1{\bm{1}}

\DeclareMathAlphabet{\mathsfit}{\encodingdefault}{\sfdefault}{m}{sl}
\SetMathAlphabet{\mathsfit}{bold}{\encodingdefault}{\sfdefault}{bx}{n}

\usepackage{hyperref}
\usepackage{url}
\usepackage{booktabs}
\usepackage{graphicx}
\usepackage[table]{xcolor}
\usepackage[capitalise]{cleveref}
\makeatletter
\AddToHook{cmd/appendix/before}{\def\cref@section@alias{appendix}}
\makeatother

\crefname{appendix}{Appendix}{Appendices}
\Crefname{appendix}{Appendix}{Appendices}
\crefname{subappendix}{Appendix}{Appendices}
\Crefname{subappendix}{Appendix}{Appendices}
\crefname{subsubappendix}{Appendix}{Appendices}
\Crefname{subsubappendix}{Appendix}{Appendices}

\let\oldappendix\appendix
\renewcommand{\appendix}{%
  \oldappendix%
  \crefalias{section}{appendix}%
  \crefalias{subsection}{subappendix}%
  \crefalias{subsubsection}{subsubappendix}%
}

\crefformat{equation}{Eq.~#2#1#3}
\usepackage{wrapfig}
\usepackage{subcaption}

\usepackage[most]{tcolorbox}
\usepackage{verbatim}
\usepackage{listings}
\usepackage[frozencache,cachedir=.]{minted}
\usepackage{fancyvrb}
\fvset{fontfamily=\rmdefault}
\NewDocumentCommand{\VarField}{m}{\texttt{\{#1\}}}
\DefineVerbatimEnvironment{PromptVerbatim}
  {Verbatim}
  {commandchars=\\\{\}, breaklines=true}
\lstnewenvironment{myverbatim}[1][]{%
  \lstset{
    basicstyle=\ttfamily,
    frame=tb,
    #1
  }%
}{}
\BeforeBeginEnvironment{myverbatim}{%
\refstepcounter{myverb}%
}

\definecolor{darkblue}{rgb}{0, 0, 0.5}
\hypersetup{colorlinks=true, citecolor=darkblue, linkcolor=darkblue, urlcolor=darkblue}

\newcommand{\myparagraph}[1]{\noindent\textbf{#1\hspace{0.5em}}}

\newcommand{\nocomment}{}

\ifx\nocomment\undefined
\else
\fi

\definecolor{train-color}{HTML}{D81B60}
\definecolor{eval-color}{HTML}{1E88E5}

\newcommand{\C}[1]{%
  \ifboolexpr{
    test {\ifstrequal{#1}{1}}
    or
    test {\ifstrequal{#1}{2}}
  }
    {\ensuremath{\textcolor{train-color}{\text{C}_{\text{#1}}}}}
    {\ensuremath{\textcolor{eval-color}{\text{C}_{\text{#1}}}}}%
}

\title{Evaluating Persistent Calibration under Evolving Model Knowledge}

\author{Victor Wang\textsuperscript{1} \quad Thomas Hofweber\textsuperscript{2} \quad Mohit Bansal\textsuperscript{2} \quad Elias Stengel-Eskin\textsuperscript{1} \\[2mm] \textsuperscript{1}University of Texas at Austin \quad \textsuperscript{2}University of North Carolina at Chapel Hill
}

\iclrfinalcopy 
\begin{document}

\maketitle

\begin{abstract}
As AI systems move from static repositories to agents that are capable of continual adaptation and learning, maintaining their trustworthiness means equipping the models backing them with the ability to produce confidence estimates that dynamically reflect their changing skills and knowledge.
We introduce the problem of \textit{persistent calibration}, which requires a confidence estimator to faithfully reflect the knowledge contained in a model as that knowledge changes, without recurring supervision.
We operationalize this by examining persistent calibration across checkpoints of open models, asking whether confidence estimators trained on earlier checkpoints can generalize to later ones.
Specifically, we aim to shed light on whether confidence is dependent on knowledge, a question with implications for the reliability of confidence estimates.
To measure this relationship, we define and evaluate calibration on \textit{knowledge contrast sets}: subsets containing questions that one checkpoint answers correctly and another checkpoint answers incorrectly, reflecting a change in knowledge. 
We show that both inference-time and fine-tuning methods fall short on contrast-set calibration compared to oracle methods trained on future checkpoints, even for methods that are well-calibrated on the full dataset.
We provide evidence for the hypothesis that persistent calibration is challenging because there is a vast space of possible confidence functions that are well-calibrated on a given checkpoint, out of which only some rely on meta-knowledge features that would generalize to other checkpoints. 
Towards improving contrast-set calibration, we show that multi-checkpoint training helps, suggesting an avenue for identifying confidence features that remain robust across changing knowledge.\footnote{Code: \url{https://github.com/victorwang37/persistent-calibration}}
\end{abstract}

\section{Introduction}

The models that back AI systems deployed in the real world must continually learn and explore \citep{ring1994continual, shi2025continual}.
For example, novel events can override past knowledge \citep{kasai2023realtime}, and in domains like scientific discovery, models must internalize and build on an evolving literature \citep{zhang2025exploring}.
Such continually learning models ought to remain aligned even as they further acquire knowledge over time \citep{bai2022constitutional, wang2023trace, wynn2024learning}.
One particularly critical alignment direction which pertains directly to their knowledge is epistemic humility:  
knowing the boundaries of one's knowledge, communicated with calibrated confidence estimates that are indicative of the probability that a generated claim is correct \citep{zhou2024relying, steyvers2025large}.
This is a meta-capability that we would hope for our models -- typically, large language models (LLMs) -- to retain as their knowledge evolves, as it is critical in developing trustworthy AI systems.
The goal of preserving calibration as a model's knowledge evolves motivates our key research question about the relationship between continual learning and confidence calibration: 
\textbf{\emph{Can a confidence estimator track a model's knowledge such that, after a change in knowledge, the new confidence is calibrated with respect to the new knowledge?}}

\begin{figure}[t]
    \centering
    \includegraphics[width=\linewidth]{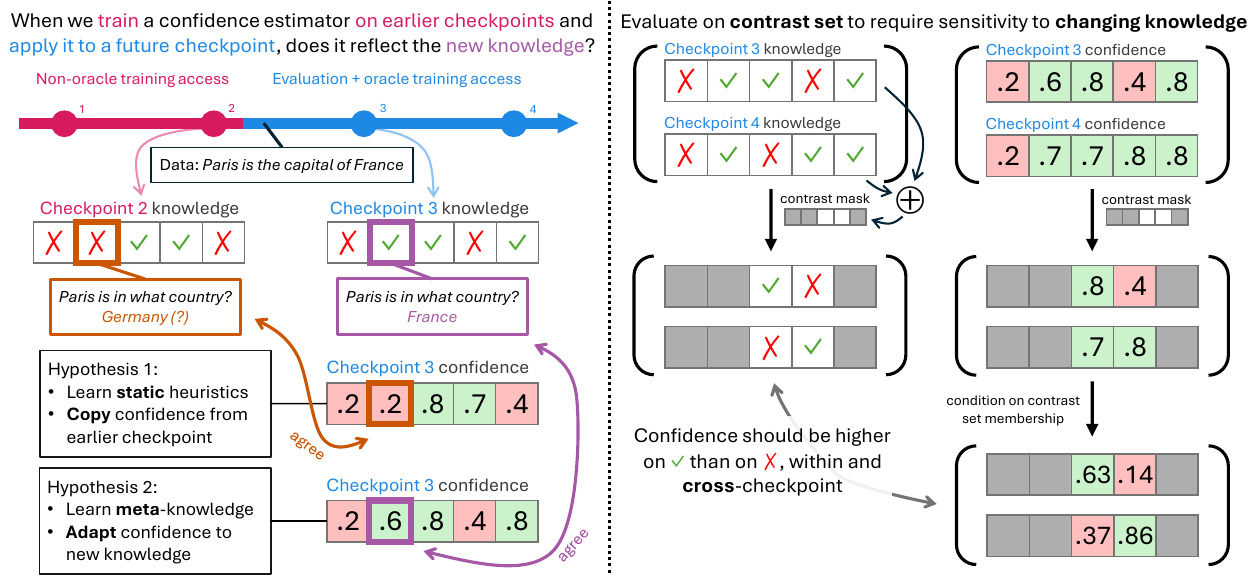}
    \caption{\textbf{(Left)} Our setup for evaluating persistent calibration.
    During training, the model encounters the training data instance \textit{``Paris is the capital of France''}, causing a change in knowledge between checkpoints \C2 and \C3.
    A confidence estimator that is faithful to the model's evolving knowledge (Hypothesis 2) should produce a higher confidence on the checkpoint \C3 with increased knowledge; however, an estimator could instead capture Hypothesis 1, which performs equally well on the training distribution \C2 but fails to generalize to \C3. 
    \textbf{(Right)} To measure the dependence of confidence on knowledge, we evaluate calibration on a contrast set consisting of questions that one checkpoint answers correctly and another checkpoint answers incorrectly, testing whether a change in knowledge is accompanied by a change in confidence.}
    \label{fig:main}
\end{figure}

To study this question, we introduce the notion of \textit{persistent calibration}, which requires a model to produce confidence estimates calibrated to its dynamically evolving knowledge at any given point in its training trajectory, without requiring repeated re-calibration.
As a testbed for persistent calibration, we use checkpoints of open models \citep{olmo2025olmo, marin}.
We test the persistent calibration of various confidence estimation methods: for training-based methods, we grant them access to earlier checkpoints and evaluate them on the later evaluation checkpoints, testing whether the learned features transfer to future, unseen checkpoints in which the model's knowledge has evolved (\cref{fig:main}, left); for inference-time methods (e.g.\ self-consistency), we directly apply them to the evaluation checkpoints. 
However, this evaluation is confounded by the fact that a given model's predictions are highly correlated across checkpoints (e.g.\ common facts are learned first and tend not to be forgotten, while obscure ones are never learned \citep{mallen2023not}), so strong future-checkpoint calibration can be at least partially explained by a high inter-checkpoint correlation in answer correctness rather than faithfully reflecting evolving knowledge.

With a view to controlling for this confounder, we introduce the notion of the \textit{knowledge contrast set}, which consists of questions where answer correctness differs by checkpoint.
This decorrelates the checkpoints and explicitly requires sensitivity to a given checkpoint's knowledge set, since a single confidence value cannot capture why one checkpoint answers correctly while another checkpoint answers incorrectly (\cref{fig:main}, right).
Moreover, we argue that contrast sets help disentangle whether a method reflects meta-knowledge or static heuristics.
Prior work has suggested that meta-knowledge (knowledge about knowledge, i.e., a model knowing it does/does not know something) may not be necessary to achieve confidence calibration, as it suffices to use heuristics such as question difficulty \citep{zhu2026a, xiao2026generalized}.
We argue that such heuristics are unsuitable for persistent calibration and would have poor contrast-set calibration, since, for example, a question that is initially difficult may become easy after more training.
On the other hand, a method that identifies meta-knowledge features ought to have better contrast-set calibration: if it can identify fact-independent features that represent meta-knowledge of whether a fact is known or unknown, it should be able to adapt when the model's knowledge about the fact changes.

We hypothesize that learning a confidence estimator capable of persistent calibration is challenging because of the vast space of confidence functions that are well-calibrated on a given checkpoint, out of which only some use meta-knowledge in a way that generalizes to other checkpoints; this intuition is illustrated in \cref{fig:main} (left).
Understanding whether a confidence estimator reflects meta-knowledge has practical implications for intervening on a model's knowledge via continual learning.
When a model is trained over long time spans and encounters new topics or domains, it can undergo drastic changes in knowledge, making earlier performance no longer representative of current performance \citep{gururangan2020don, wang2025parameter, periodiclabs2026nature}, yet we would still like to know how trustworthy the current model is.
At a more granular level, we may seek to invest more resources into a target problem when uncertain, such as with retrieval \citep{asai2024self}, tool use \citep{wang2026position}, or reasoning \citep{guo2025deepseek}.
In such cases of adaptive effort, confidence estimates should reflect whether a given attempt has improved the probability of success, or whether the model should spend more effort or abstain.
An estimator that is insensitive to the expansion or reduction of a model's capabilities is bound to become obsolete as models continue to grow more capable.

Using our cross-checkpoint knowledge contrast sets on TriviaQA, Jeopardy, and BioASQ, we evaluate confidence estimation methods from various families, including fine-tuning \citep{kapoor2024large, li2026conftuner} and inference-time approaches \citep{kuhn2023semantic, manakul2023selfcheckgpt}.
Since fine-tuning methods typically operate on a single snapshot of a model, it is plausible that they learn heuristics that work specifically on the current model and do not robustly generalize to the knowledge sets of future model versions. 
Inference-time (training-free) methods like self-consistency sidestep the bias of checkpoint-specific feature representations by virtue of treating models as black boxes, making them appealing as a potential solution for persistent calibration.
However, it is unclear whether inference-time methods access fine-grained meta-knowledge as opposed to shallow cues \citep{singh2026can}, and forgoing training typically means worse calibration than fine-tuning methods on in-distribution data.
To quantify the room for improvement of these methods, we compare them to oracle methods that are allowed to train on the evaluation checkpoints.
We find that existing methods obtain limited contrast-set calibration despite having strong full-set calibration.
For example, for Marin 8B on Jeopardy, self-consistency matches oracle training on full-set AUC (0.886 vs.\ 0.890), despite being far behind on contrast-set AUC (0.760 vs.\ 0.825).

In both oracle and non-oracle settings, we find that contrast-set calibration improves when we introduce multi-checkpoint training.
In particular, multi-checkpoint training in the oracle setting outperforms respective-checkpoint training (i.e.\ training one adapter per checkpoint), revealing that training a single adapter to learn confidence features that work on multiple checkpoints has a synergistic effect in identifying features that are sensitive to differences in knowledge across checkpoints.
We further analyze the heterogeneity of confidence estimators that have the same performance on a given checkpoint, showing that such confidence estimators can have vastly different behaviors from each other on other checkpoints.
This heterogeneity presents a challenge for identifying robust meta-knowledge features that are conducive to persistent calibration.

\section{Persistent Calibration}
\label{sec:persistent-calibration}

We introduce persistent calibration as a criterion for confidence estimators: they should faithfully reflect a model's knowledge as it changes, as this is critical for trustworthy, continually learning models.
To tractably evaluate persistent calibration, we leverage the checkpoint trajectories of open models trained on trillions of tokens from diverse corpora \citep{olmo2025olmo, marin}, offering a large-scale setup to study the evolution of knowledge over time.
We evaluate both training-based methods and inference-time methods.
For training-based methods, as shown in \cref{fig:main} (left), we grant them access to earlier checkpoints and evaluate them on the later evaluation checkpoints.
For inference-time methods, we directly apply them to the evaluation checkpoints.
By default, for both training and evaluation, the checkpoint that generated an answer is the same checkpoint that produces a confidence estimate for the answer.
We describe generating answers for our experiments in \cref{sec:confidence-estimation-methods}.

\paragraph{Knowledge Contrast Sets.} Our research question concerns whether a model's current confidence estimates are faithful to its current knowledge as opposed to the knowledge of another model version.
Due to the non-random overlap between different checkpoints' knowledge sets (e.g.\ common facts are known by most checkpoints \citep{chang2024large}), confidence estimates that are well-calibrated with respect to one checkpoint's knowledge may also be well-calibrated with respect to another checkpoint's knowledge, so strong calibration is insufficient to show faithfulness.
To address this confound, we evaluate calibration on a \textit{knowledge contrast set}, which, for a given pair of checkpoints, consists of questions that one checkpoint answers correctly and the other checkpoint answers incorrectly, as shown in \cref{fig:main} (right).
This construction means that learning to predict confidence based on checkpoint-agnostic question features (e.g.\ topic or difficulty) may improve full-set calibration but would not improve contrast-set calibration, because the predicted confidence should not change (as the question is the same across checkpoints) and thus the predicted confidence would only be consistent with one checkpoint's correctness but not the other.
Faithful confidence estimates should be higher on the checkpoint that predicts the correct answer, similar to how confidence should tend to be higher for correct answers than for incorrect answers within a checkpoint.

\paragraph{Conditioning on Contrast Set Membership.} Confidences that are globally calibrated may appear miscalibrated within the contrast subset due to selection bias: the contrast set accuracy (pooled over the two checkpoints) is always 50\%, regardless of the global accuracies.
To correct for this, given a contrast-set question where checkpoints 1 and 2 have confidences $p_1$ and $p_2$ in their respective answers, we normalize them to $p_1 (1 - p_2) / Z$ and $p_2 (1 - p_1) / Z$, where $Z = p_1 (1 - p_2) + p_2 (1 - p_1)$.
This restores contrast-set calibration if $p_1, p_2$ are calibrated and the correctness outcomes are conditionally independent given $p_1, p_2$.
We discuss this decision in \cref{app:condition-conset}, and provide empirical support in \cref{app:learn-condition-conset} by showing that our normalization function and a fitted normalization function produce similar results.
Moreover, the metrics $\Delta_0, \Delta^b_0$ (defined below) simply compare whether $p_1 > p_2$ (unaffected by normalization), and our results show that their trends track with the other metrics, which depend on normalization.

\paragraph{Evaluation Metrics.} For both the full set and the contrast set, we report standard metrics for the discriminability and calibration of confidence estimates: area under the receiver operating characteristic curve \citep[AUC;][]{hanley1982meaning}, Brier score \citep[BS;][]{brier1950verification}, and Expected Calibration Error \citep[ECE;][]{naeini2015obtaining},
all on a $[0, 1]$ scale.
See \cref{app:eval-metrics} for descriptions.
We compute ECE with SmoothECE for its continuous and hyperparameter-free properties \citep{blasiok2024smooth}.
For the contrast set $\mathcal{D}_{\rm con}$, we also report metrics that compare, for a given question $q^i$, the confidence $p^i_+$ on the correct answer and the confidence $p^i_-$ on the incorrect answer: $\Delta = \frac{1}{|\mathcal{D}_{\rm con}|} \sum_{i \in \mathcal{D}_{\rm con}} p^i_+ - p^i_-$ and $\Delta_0 = \frac{1}{|\mathcal{D}_{\rm con}|} \sum_{i \in \mathcal{D}_{\rm con}} \operatorname{sgn}(p^i_+ - p^i_-)$.
To control for performance coming from the prior that the later checkpoint tends to be correct, we also report class-balanced variants denoted $\Delta^b, \Delta^b_0$, where the two classes $\mathcal{D}_{\rm imp}, \mathcal{D}_{\rm reg}$ are improvement (i.e., the later checkpoint is correct) and regression (i.e., the later checkpoint is incorrect): $\Delta^b = \frac12 \left( \frac{1}{|\mathcal{D}_{\rm imp}|} \sum_{i \in \mathcal{D}_{\rm imp}} (p^i_+ - p^i_-) + \frac{1}{|\mathcal{D}_{\rm reg}|} \sum_{i \in \mathcal{D}_{\rm reg}} (p^i_+ - p^i_-) \right)$
and analogously for $\Delta^b_0$.
These $\Delta$ metrics are on a scale of $[-1, 1]$, e.g.\ random guessing gets $\Delta_0 = 0$, while always knowing which of the two answers is correct gets $\Delta_0 = 1$.
We evaluate on multiple future checkpoints for a robust evaluation.
See \cref{app:agg-over-ckpts} for how we aggregate over them when reporting metrics.

\section{Experiments and Results}
\label{sec:experiments}

\subsection{Experimental Setup}

\paragraph{Models.} As we define persistent calibration as a multi-checkpoint property, we use open models that release multiple checkpoints.
We evaluate on the pre-training checkpoints of Olmo 3 7B and 32B \citep{olmo2025olmo} and Marin 8B \citep{marin}.
For Olmo 3 models, we use checkpoints at specified percentages of pre-training progress; we take [10\%, 20\%, 30\%] as non-oracle training checkpoints, and [40\%, 50\%, 90\%, 100\%] as evaluation checkpoints; we place a gap in the evaluation checkpoints to induce a substantial accuracy difference for our contrast sets.
For Marin 8B, we use all 6 available checkpoints: the first 3 for training and the last 3 for evaluation.

\paragraph{Datasets.} We evaluate on TriviaQA \citep{joshi2017triviaqa}, Jeopardy \citep{openaccessaicollective_jeopardy}, and BioASQ \citep{krithara2023bioasq}, which test factual knowledge recall.
See \cref{app:data-processing} for details on data processing.
TriviaQA and Jeopardy require diverse world knowledge, while BioASQ requires biomedical expertise.
For each of TriviaQA and Jeopardy, we use 5K questions for training and 5K questions for evaluation.
Due to the smaller size of BioASQ, we only run evaluation to test domain transfer from TriviaQA and Jeopardy.
We use 5 fixed in-context examples from the respective dataset for answer generation.
We use an LLM judge to label the correctness of a predicted answer with respect to the reference ground-truth answer \citep{wei2024measuring}; in our manual audit (\cref{app:audit-data-quality}), we confirm that LLM and human judgments have high agreement, and that data quality is high for both contrast sets and full sets.

\subsubsection{Confidence Estimation Methods}
\label{sec:confidence-estimation-methods}

We evaluate representative confidence estimation methods for persistent calibration, using both full datasets and their contrast subsets.
As a running example, we suppose a sequence of checkpoints [\C1, \C2, \C3, \C4] as in \cref{fig:main}, where [\C1, \C2] are for non-oracle training and [\C3, \C4] are for evaluation and oracle training.
To quantify the room for improvement of a confidence estimator, we compare to a confidence adapter trained on the eval checkpoints [\C3, \C4].
We refer to this as an \emph{oracle} method as it allows directly learning confidence features from [\C3, \C4], whereas non-oracle methods must learn meta-knowledge features (from [\C1, \C2], if training) that generalize to [\C3, \C4].

\paragraph{Self-Consistency (SC).} SC quantifies a model's confidence in an answer as its probability of generating it, typically estimated via sampling \citep{kuhn2023semantic}.
Pre-trained LLMs often have well-calibrated generation probabilities \citep{zhu2023calibration, luo2026your}, since they are trained to model the true data distribution with next-token prediction.
While prior work has also explored verbalizing confidence from the model \citep{tian2023just, wang2026calibrating}, this generally requires an instruction-tuned model, and -- as confirmed by early experiments -- does not perform well for pre-trained checkpoints of the kind we evaluate.
Thus, we omit zero-shot verbalized confidence.
We add two implementation details that improve SC's robustness:
rather than sampling, we run beam search (beam size 10) and weight each beam by its sequence probability to efficiently cover the distribution of short-form generations \citep{fadeeva2026don}, and
we use an off-the-shelf NLI model to group equivalent answers \citep[][see \cref{app:sc-implementation}]{kuhn2023semantic}.
We define a candidate answer as a set of NLI-grouped beams, with generation probabilities summed.
We take the top (i.e.\ highest probability) candidate answer as the model's prediction, and its probability as the SC confidence.

\paragraph{LoRA + Verbalized Confidence.} Although pre-trained models struggle to verbalize confidence zero-shot, we can fine-tune them to do so.
We use LoRA \citep{hu2022lora} for its parameter efficiency and expressiveness \citep{kapoor2024large}.
We prompt the model to verbalize its confidence on a scale of 0 to 9 and compute the confidence as a probability-weighted average over the tokens \texttt{0}, \ldots, \texttt{9} \citep{li2026conftuner}.
See \cref{app:lora-config} for the training configuration.
We introduce multi-checkpoint and multi-answer training to learn a better confidence adapter.
\textbf{Multi-checkpoint training} trains one adapter to produce calibrated confidence estimates for multiple checkpoints (using their respective predictions, by default).
This is motivated by the intuition that a confidence adapter trained on a single checkpoint \C2 may pick up on features that are specific to \C2, while multi-checkpoint training may encourage learning generalizable features.
We apply multi-checkpoint training to [\C1, \C2] (non-oracle) or [\C3, \C4] (oracle).
We use multi-checkpoint training in \cref{tab:oracle-vs-methods-abridged,tab:methods-vs-ablations-triviaqa} and ablate it in \cref{tab:multi-vs-single-ckpt-triviaqa}.
\textbf{Multi-answer training} uses multiple top candidate answers for a given question.
The motivation of this is to learn answer sensitivity, since seeing only \C2's top-1 predictions may lead the adapter to anticipate \C2's answer and ignore the candidate answer, which may differ between \C2 and [\C3, \C4].
We compute a question's loss as the average of its candidate answers' losses (same batch), weighted by their generation probabilities.
This ensures that each question's total loss weight is a constant, and that the training distribution is representative of the model's generation distribution.
We train with 2 answers by default, and we ablate this in \cref{app:ablation-multi-ckpt-multi-answer}.
We ensure controlled comparisons: for a given training seed, all methods receive the same LoRA initialization and order of questions; for multi-checkpoint training, the checkpoints share the question shuffle and the loss is averaged over checkpoints so the effective learning rate is unaffected.
Multi-checkpoint and multi-answer training can be seen as data augmentation of the curated (question, reference answer) data.

\paragraph{End-correct Baseline.} To control for the prior that accuracy increases over training, this baseline predicts a fixed confidence based just on checkpoint index.
We linearly interpolate $[0, 1]$ with the evaluation checkpoints; for three checkpoints \C3, \C4, \C5, we would get fixed confidences [0, 1/2, 1].

\paragraph{Generalized Correctness Models (GCMs).} \citet{xiao2026generalized} introduce GCMs to predict the correctness of other models' answers.
GCMs show similar or better calibration compared to using the other models themselves, challenging the premise that calibration requires meta-knowledge.
We revisit this result using our contrast sets, which are designed to require meta-knowledge.
We run the TriviaQA-trained GCM released by \citet{xiao2026generalized}, using the model-agnostic prompt containing only the question and the candidate answer.

\subsection{Results}
\label{sec:main-results}

\begin{table}[t]
\caption{SC and non-oracle training consistently fall short on contrast-set calibration compared to oracle training, even when full-set calibration is saturated.
Results in \cref{sec:experiments} are averaged over 3 seeds.
In each column, we bold the best result and underline results not significantly worse under a paired test ($\alpha = 0.05$; see \cref{app:sigtests} for tests); tables in \cref{sec:experiments} share this text styling. See \cref{app:oracle-vs-methods-unabridged} for all evaluation metrics and Jeopardy results.
}
\label{tab:oracle-vs-methods-abridged}
\vspace{-0.75em}
\begin{center}
\setlength{\tabcolsep}{8pt}
\small
\resizebox{\linewidth}{!}{%
\renewcommand{\arraystretch}{1.3}
\begin{tabular}{lccccccccc}
\toprule
& \multicolumn{3}{c}{Olmo 3 7B} & \multicolumn{3}{c}{Marin 8B} & \multicolumn{3}{c}{Olmo 3 32B} \\
\cmidrule(lr){2-4}\cmidrule(lr){5-7}\cmidrule(lr){8-10}
& \multicolumn{2}{c}{Contrast} & \multicolumn{1}{c}{Full} & \multicolumn{2}{c}{Contrast} & \multicolumn{1}{c}{Full} & \multicolumn{2}{c}{Contrast} & \multicolumn{1}{c}{Full} \\
\cmidrule(lr){2-3}\cmidrule(lr){4-4}\cmidrule(lr){5-6}\cmidrule(lr){7-7}\cmidrule(lr){8-9}\cmidrule(lr){10-10}
Method & $\Delta_0^b$ $\scriptstyle\uparrow$ & AUC $\scriptstyle\uparrow$ & AUC $\scriptstyle\uparrow$ & $\Delta_0^b$ $\scriptstyle\uparrow$ & AUC $\scriptstyle\uparrow$ & AUC $\scriptstyle\uparrow$ & $\Delta_0^b$ $\scriptstyle\uparrow$ & AUC $\scriptstyle\uparrow$ & AUC $\scriptstyle\uparrow$ \\
\midrule
& \multicolumn{9}{c}{TriviaQA} \\
\cmidrule(lr){2-10}
End-correct baseline & 0.000 & 0.627 & 0.530 & 0.000 & 0.783 & 0.547 & 0.000 & 0.712 & 0.548 \\
Self-consistency & 0.297 & 0.733 & \underline{0.884} & 0.263 & \underline{0.823} & \underline{0.897} & \underline{0.362} & \underline{0.808} & \textbf{0.903} \\
Non-oracle training & 0.333 & 0.748 & 0.880 & 0.227 & 0.806 & 0.834 & 0.285 & 0.688 & 0.860 \\
Oracle training & \textbf{0.379} & \textbf{0.795} & \textbf{0.890} & \textbf{0.349} & \textbf{0.840} & \textbf{0.901} & \textbf{0.384} & \textbf{0.827} & \underline{0.898} \\
\midrule
& \multicolumn{9}{c}{Jeopardy} \\
\cmidrule(lr){2-10}
End-correct baseline & 0.000 & 0.633 & 0.540 & 0.000 & 0.735 & 0.549 & 0.000 & 0.653 & 0.549 \\
Self-consistency & 0.333 & 0.741 & 0.873 & 0.331 & 0.760 & \underline{0.886} & 0.320 & 0.754 & \underline{0.893} \\
Non-oracle training & 0.476 & 0.821 & 0.871 & 0.397 & 0.759 & 0.834 & 0.399 & 0.757 & 0.860 \\
Oracle training & \textbf{0.506} & \textbf{0.855} & \textbf{0.888} & \textbf{0.459} & \textbf{0.825} & \textbf{0.890} & \textbf{0.475} & \textbf{0.840} & \textbf{0.894} \\
\bottomrule
\end{tabular}
}
\end{center}
\end{table}

\myparagraph{Confidence Estimators Struggle on Contrast Sets.} \cref{tab:oracle-vs-methods-abridged} shows that SC and non-oracle training consistently have worse contrast-set calibration than oracle training, showing room for improvement.
We highlight that even when SC matches oracle training on full-set AUC (e.g., SC 0.886 vs.\ oracle 0.890 on Marin 8B, Jeopardy), SC is often far behind on contrast-set AUC (SC 0.760 vs.\ oracle 0.825).
Thus, full-set calibration does not reliably predict contrast-set calibration, establishing the latter as a distinct criterion for persistent calibration.
Moreover, this supports our hypothesis that there are various confidence estimators that have the same aggregate (i.e.\ full-set) performance and yet substantially different behavior on knowledge changes featured in the contrast set (other full-set metrics do not explain this difference either, \cref{app:oracle-vs-methods-unabridged}), making it hard to learn a confidence estimator that is capable of persistent calibration.
We discuss this further in \cref{sec:analysis}.

\myparagraph{Domain Transfer.} In \cref{app:domain-transfer}, we find that the performance gain of oracle training over non-oracle training observed in-domain (\cref{tab:oracle-vs-methods-abridged}) largely disappears when transferring to BioASQ, as both drop to similarly low performances.
This clarifies the scope of the oracle's advantage in our experiments as being limited to the training domain, consistent with prior work on the domain-specificity of correctness signals \citep{sky2024androids}.

\begin{table}[t]
\caption{Comparing SC and non-oracle training with ablations controlling for knowledge already present in an earlier checkpoint. Beating these ablations is necessary for a nontrivial demonstration of persistent calibration. 
SC generally beats its ablations, while non-oracle training has mixed results. 
Results on TriviaQA; see \cref{app:oracle-vs-methods-unabridged} for all metrics and Jeopardy results.}
\label{tab:methods-vs-ablations-triviaqa}
\vspace{-0.75em}
\begin{center}
\setlength{\tabcolsep}{8pt}
\small
\resizebox{\linewidth}{!}{%
\renewcommand{\arraystretch}{1.3}
\begin{tabular}{lccccccccc}
\toprule
& \multicolumn{3}{c}{Olmo 3 7B} & \multicolumn{3}{c}{Marin 8B} & \multicolumn{3}{c}{Olmo 3 32B} \\
\cmidrule(lr){2-4}\cmidrule(lr){5-7}\cmidrule(lr){8-10}
& \multicolumn{2}{c}{Contrast} & \multicolumn{1}{c}{Full} & \multicolumn{2}{c}{Contrast} & \multicolumn{1}{c}{Full} & \multicolumn{2}{c}{Contrast} & \multicolumn{1}{c}{Full} \\
\cmidrule(lr){2-3}\cmidrule(lr){4-4}\cmidrule(lr){5-6}\cmidrule(lr){7-7}\cmidrule(lr){8-9}\cmidrule(lr){10-10}
Method & $\Delta_0^b$ $\scriptstyle\uparrow$ & AUC $\scriptstyle\uparrow$ & AUC $\scriptstyle\uparrow$ & $\Delta_0^b$ $\scriptstyle\uparrow$ & AUC $\scriptstyle\uparrow$ & AUC $\scriptstyle\uparrow$ & $\Delta_0^b$ $\scriptstyle\uparrow$ & AUC $\scriptstyle\uparrow$ & AUC $\scriptstyle\uparrow$ \\
\midrule
Self-consistency & \textbf{0.297} & \textbf{0.733} & \textbf{0.884} & \underline{0.263} & \textbf{0.823} & \textbf{0.897} & \textbf{0.362} & \textbf{0.808} & \textbf{0.903} \\
\,\raisebox{0.5ex}{\(\llcorner\)} Surrogate & 0.211 & 0.637 & 0.859 & \textbf{0.277} & 0.695 & 0.871 & 0.288 & 0.688 & 0.883 \\
\,\raisebox{0.5ex}{\(\llcorner\)} Copy & 0.000 & 0.500 & 0.832 & 0.000 & 0.500 & 0.829 & 0.000 & 0.500 & 0.846 \\
\midrule
\addlinespace[3pt]
Non-oracle training & \textbf{0.333} & \textbf{0.748} & \textbf{0.880} & 0.227 & \textbf{0.806} & 0.834 & \underline{0.285} & 0.688 & \textbf{0.860} \\
\,\raisebox{0.5ex}{\(\llcorner\)} Surrogate & \underline{0.318} & 0.729 & 0.869 & \textbf{0.318} & 0.748 & \textbf{0.866} & \textbf{0.290} & \textbf{0.716} & \underline{0.858} \\
\,\raisebox{0.5ex}{\(\llcorner\)} Copy & 0.000 & 0.500 & 0.839 & 0.000 & 0.500 & 0.830 & 0.000 & 0.500 & 0.832 \\
\bottomrule
\end{tabular}
}
\end{center}
\end{table}

\myparagraph{Ablations to Test Nontrivial Transfer.} Although we have seen that SC and non-oracle training have imperfect contrast-set calibration, it appears nontrivial, outperforming the end-correct baseline in almost all cases (\cref{tab:oracle-vs-methods-abridged}).
We examine this further with two ablations.
The \textbf{surrogate} ablation uses \C2 to estimate confidence on [\C3, \C4]'s answers.
This ablation isolates the performance that can be achieved with the knowledge already contained in \C2, without needing to inspect the knowledge of [\C3, \C4].
The \textbf{copy} ablation uses \C2 to estimate confidence on its own answers and simply reuses these confidences for [\C3, \C4] answering the same questions.
This ablation is oblivious to [\C3, \C4] and their answers, relying solely on the inter-checkpoint correlation in correctness.
A confidence estimator must outperform these two ablations for a convincing demonstration of faithfulness to a given checkpoint's (\C3 or \C4) changed knowledge that cannot be explained by prior knowledge.

\myparagraph{Results.} \cref{tab:methods-vs-ablations-triviaqa} shows that SC generally outperforms its ablations, while non-oracle training has mixed results.
The faithfulness of SC is perhaps expected given that it treats models as black boxes, avoiding checkpoint-specific features.
The mixed results for non-oracle training suggest that a confidence adapter trained on a set of checkpoints [\C1, \C2] sometimes but unreliably learns features that generalize beyond to [\C3, \C4].
Lastly, we observe that the $\rm copy$ ablations -- with no awareness of [\C3, \C4] or their answers -- achieve strong full-set AUC (consistently over 0.8) but a trivial contrast-set AUC of 0.5.
This highlights that full-set performance is a poor reflection of the faithfulness of a confidence estimator and is inflated by inter-checkpoint correlation of correctness.

\begin{table}[t]
\caption{Multi-checkpoint training improves contrast-set and full-set calibration. Results on TriviaQA; see \cref{app:ablation-multi-ckpt-multi-answer} for Jeopardy.
}
\vspace{-0.75em}
\label{tab:multi-vs-single-ckpt-triviaqa}
\begin{center}
\setlength{\tabcolsep}{8pt}
\small
\resizebox{\linewidth}{!}{%
\renewcommand{\arraystretch}{1.3}
\begin{tabular}{lccccccccc}
\toprule
& \multicolumn{3}{c}{Olmo 3 7B} & \multicolumn{3}{c}{Marin 8B} & \multicolumn{3}{c}{Olmo 3 32B} \\
\cmidrule(lr){2-4}\cmidrule(lr){5-7}\cmidrule(lr){8-10}
& \multicolumn{2}{c}{Contrast} & \multicolumn{1}{c}{Full} & \multicolumn{2}{c}{Contrast} & \multicolumn{1}{c}{Full} & \multicolumn{2}{c}{Contrast} & \multicolumn{1}{c}{Full} \\
\cmidrule(lr){2-3}\cmidrule(lr){4-4}\cmidrule(lr){5-6}\cmidrule(lr){7-7}\cmidrule(lr){8-9}\cmidrule(lr){10-10}
Method & $\Delta_0^b$ $\scriptstyle\uparrow$ & AUC $\scriptstyle\uparrow$ & AUC $\scriptstyle\uparrow$ & $\Delta_0^b$ $\scriptstyle\uparrow$ & AUC $\scriptstyle\uparrow$ & AUC $\scriptstyle\uparrow$ & $\Delta_0^b$ $\scriptstyle\uparrow$ & AUC $\scriptstyle\uparrow$ & AUC $\scriptstyle\uparrow$ \\
\midrule
\multicolumn{1}{l}{\:\textit{Non-oracle}} & \multicolumn{9}{c}{} \\
\cmidrule(lr){1-1}
Multi-ckpt & \textbf{0.333} & \underline{0.748} & \textbf{0.880} & \textbf{0.227} & \textbf{0.806} & \textbf{0.834} & \textbf{0.285} & \textbf{0.688} & \textbf{0.860} \\
Single-ckpt & \underline{0.332} & \textbf{0.750} & 0.876 & \underline{0.221} & 0.642 & 0.783 & 0.251 & 0.673 & 0.849 \\
\midrule
\addlinespace[3pt]
\multicolumn{1}{l}{\:\textit{Oracle}} & \multicolumn{9}{c}{} \\
\cmidrule(lr){1-1}
Multi-ckpt & \textbf{0.379} & \textbf{0.795} & \textbf{0.890} & \textbf{0.349} & \textbf{0.840} & \underline{0.901} & \textbf{0.384} & \textbf{0.827} & \textbf{0.898} \\
Respective-ckpt & 0.351 & 0.778 & 0.884 & \underline{0.330} & \underline{0.837} & \textbf{0.901} & 0.352 & 0.812 & 0.892 \\
\bottomrule
\end{tabular}
}
\end{center}
\end{table}

\myparagraph{Multi-Checkpoint Training Improves Persistent Calibration.} In \cref{tab:multi-vs-single-ckpt-triviaqa}, we ablate multi-checkpoint training, which has been used so far.
The baseline in the non-oracle setting is \textbf{single-checkpoint training} on \C2. 
The baseline in the oracle setting is \textbf{respective-checkpoint training}, which trains one adapter for each of [\C3, \C4] (using the respective checkpoint's data), thus holding constant the total training data; at inference, the respective adapter is used.
\cref{tab:multi-vs-single-ckpt-triviaqa} shows that multi-checkpoint  consistently matches or exceeds the baseline, in both oracle and non-oracle settings.

\begin{table}[t]
\caption{Ablating factors contributing to the gains of multi-checkpoint training (oracle). 
$\rm Single\text{-}checkpoint\ data$ uses only one checkpoint's answers for multi-checkpoint training.
$\rm Pooled\ data$ trains each checkpoint's adapter on the same data as (unablated) multi-checkpoint training.
Multi-checkpoint training beats both, meaning that both the training data and multiple checkpoint views contribute. Results on TriviaQA, Olmo 3 7B; see \cref{app:ablation-multi-ckpt-multi-answer} for Jeopardy.}
\label{tab:multi-ckpt-vs-ablations-olmo7b-triviaqa}
\vspace{-0.75em}
\begin{center}
\setlength{\tabcolsep}{8pt}
\small
\resizebox{\linewidth}{!}{%
\renewcommand{\arraystretch}{1.3}
\begin{tabular}{lcccccccccc}
\toprule
& \multicolumn{7}{c}{Contrast} & \multicolumn{3}{c}{Full} \\
\cmidrule(lr){2-8}\cmidrule(lr){9-11}
Method & $\Delta_0^b$ $\scriptstyle\uparrow$ & $\Delta_0$ $\scriptstyle\uparrow$ & $\Delta^b$ $\scriptstyle\uparrow$ & $\Delta$ $\scriptstyle\uparrow$ & AUC $\scriptstyle\uparrow$ & BS $\scriptstyle\downarrow$ & ECE $\scriptstyle\downarrow$ & AUC $\scriptstyle\uparrow$ & BS $\scriptstyle\downarrow$ & ECE $\scriptstyle\downarrow$ \\
\midrule
Multi-ckpt & \textbf{0.379} & \textbf{0.424} & \textbf{0.255} & \textbf{0.285} & \textbf{0.795} & \textbf{0.185} & \textbf{0.061} & \textbf{0.890} & \textbf{0.131} & \underline{0.032} \\
\,\raisebox{0.5ex}{\(\llcorner\)} Single-ckpt data & \underline{0.373} & 0.368 & \underline{0.252} & 0.258 & 0.768 & 0.197 & 0.096 & 0.887 & 0.134 & 0.044 \\
Resp-ckpt & 0.351 & 0.396 & 0.241 & 0.274 & 0.778 & 0.193 & 0.070 & 0.884 & 0.133 & \textbf{0.030} \\
\,\raisebox{0.5ex}{\(\llcorner\)} Pooled data & 0.360 & 0.357 & 0.248 & 0.251 & 0.757 & 0.203 & 0.104 & 0.887 & 0.132 & 0.035 \\
\bottomrule
\end{tabular}
}
\end{center}
\end{table}

\myparagraph{Ablation of Multi-Checkpoint Training Advantages.} We analyze the potential factors contributing to the gains of multi-checkpoint training: (1) more data and (2) viewing multiple checkpoints' weights.
To remove (1), the \textbf{single-checkpoint data} ablation of multi-checkpoint training uses \C3's data for both \C3 and \C4.
This limits the training data, while still giving one adapter access to multiple checkpoints' weights, encouraging it to learn checkpoint-general features.
To remove (2), the \textbf{pooled data} ablation performs respective-checkpoint training, but trains each adapter on the data pooled over [\C3, \C4].
This allows each adapter to see the same data as the multi-checkpoint training adapter, while still only allowing each adapter to see the feature representations of its own checkpoint (\C3 or \C4).
\cref{tab:multi-ckpt-vs-ablations-olmo7b-triviaqa} shows that multi-checkpoint training outperforms both ablations, meaning that both the training data and the multiple checkpoint views contribute to its performance.
Between the two ablations, $\rm single\text{-}checkpoint\ data$ performs better (and more clearly so on Jeopardy, \cref{app:ablation-multi-ckpt-multi-answer}), suggesting that access to multiple checkpoints' weights is most responsible for the gains of multi-checkpoint training.

\begin{wrapfigure}{r}{0.48\textwidth}
\vspace{-0.35em}
    \centering
    \includegraphics[width=0.48\textwidth]{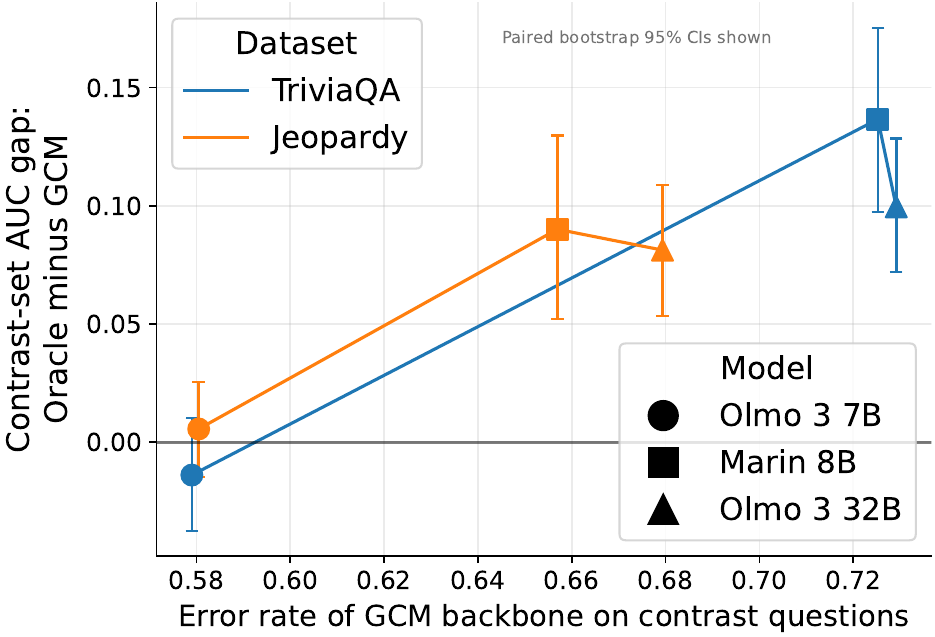}
    \vspace{-1.5em}
    \caption{GCM's headroom on contrast-set AUC grows for larger models where the GCM backbone (Qwen3-8B) knows the answer less often.
    Thus, a static external verifier does not adapt to a model's capability, motivating meta-knowledge for persistent calibration.
    }
    \vspace{-4em}
    \label{fig:gcm}
\end{wrapfigure}

\myparagraph{GCMs Do Not Satisfy Persistent Calibration.} \cref{fig:gcm} plots GCM's headroom on contrast-set AUC relative to oracle training.
Although GCM matches oracle performance on Olmo 3 7B,
with a gap close to 0, we attribute this to the GCM knowing the correct answers to many contrast set questions, enabling it to distinguish correct and incorrect answers without needing access to the internal states of the checkpoints that generated the answers.
Indeed, for the larger models (Marin 8B and Olmo 3 32B), the GCM backbone model (Qwen3-8B) has an increased error rate on contrast-set questions, and its gap from the oracle grows substantially.
Thus, we argue that while an external verifier can achieve strong calibration on the predictions of a less or similarly capable model, this ability does not transfer to more capable models, which means that a static external verifier cannot satisfy persistent calibration for continually improving models.

\begin{figure}[t]
    \centering
    \includegraphics[width=0.9\linewidth]{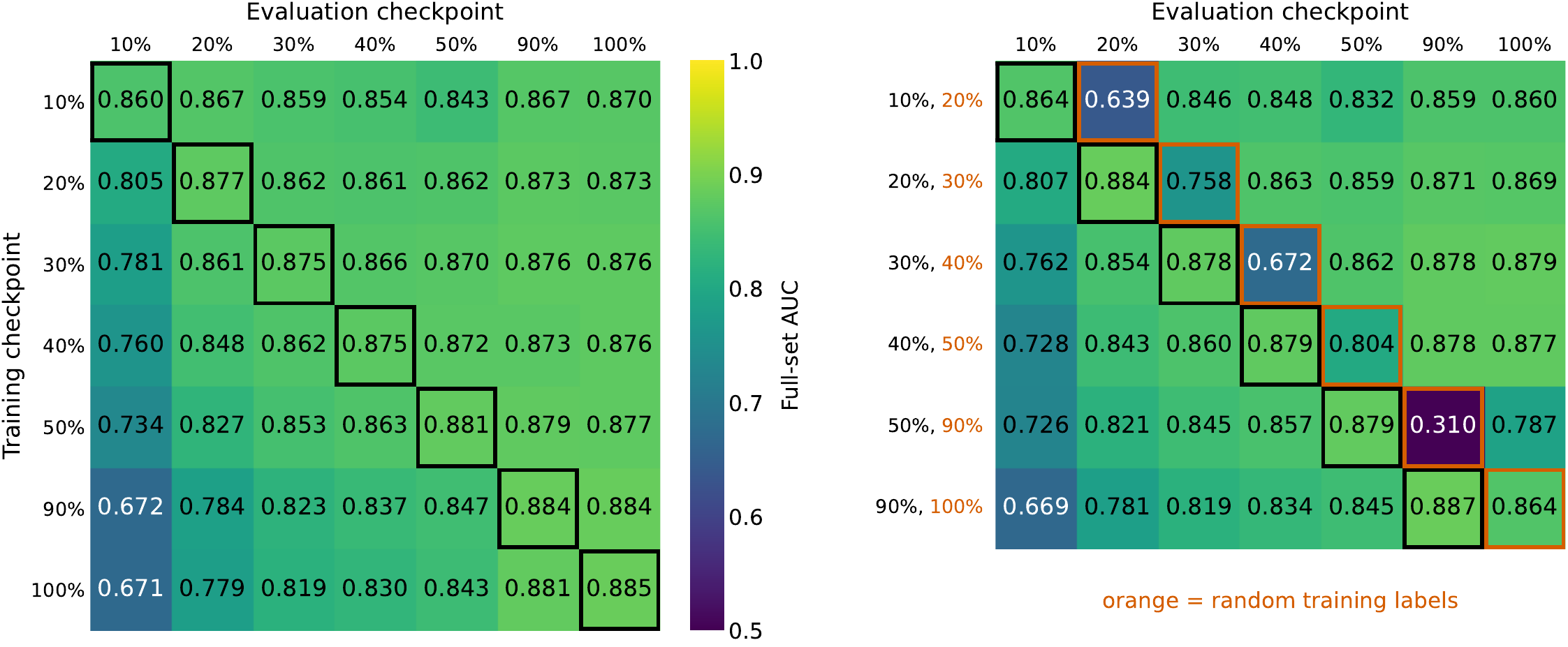}
    \caption{\textbf{(Left)} Transfer between pairs of training and eval checkpoints (Olmo 3 7B, TriviaQA; Jeopardy in \cref{app:transfer-asymmetry}). Forward transfer (top right) is strong, while backward transfer (bottom left) is weak.
    \textbf{(Right)} Transfer of multi-checkpoint training, where the second checkpoint gets random labels. The second-checkpoint AUC generally plummets, at no penalty to the first checkpoint.
    }
    \label{fig:triviaqa-transfer}
\end{figure}

\section{Analysis}
\label{sec:analysis}

The failure of non-oracle training to faithfully track knowledge (\cref{sec:experiments}) may be due to the large hypothesis space of confidence functions that are well-calibrated on a training checkpoint, only a small subset of which depend sufficiently on the meta-knowledge required for transfer.
Here, we consider a relaxed notion of transfer on the full set rather than the contrast set, given the positive full-set transfer result in \cref{sec:experiments}.
We aim to better understand the factors enabling this transfer, which may shed light on the feasibility of persistent calibration and the challenges preventing it.

\myparagraph{Transfer Asymmetry.} \cref{tab:multi-vs-single-ckpt-triviaqa} found that an adapter trained on \C2 or [\C1, \C2] is usable on \C3 or \C4, which has been trained on far more data and contains very different weights (e.g.\ the cosine similarity between the flattened weights of Olmo 3 7B 10\% and 100\% is 0.37).
To dissect this phenomenon, we compare forward transfer (i.e.\ our persistent calibration setup) with backward transfer, where an adapter is trained on \C3 or \C4 and tested on \C1 or \C2.
For efficiency in this analysis, we use single-answer training on Olmo 3 7B and one seed.
\cref{fig:triviaqa-transfer} (left) reveals that although forward transfer is strong (e.g., AUC 0.870 on 10\% $\rightarrow$ 100\%), the reverse direction is much worse (0.671 for 100\% $\to$ 10\%); this trend generally holds across checkpoints. 
The confidence adapters trained on 10\% and 100\% have nearly equal AUCs on 100\% (0.870 vs.\ 0.885) but very different AUCs on 10\% (0.860 vs.\ 0.671) -- despite matched performance on one checkpoint, two confidence adapters can implement confidence estimation in functionally different ways, as evidenced by their divergent AUCs on another checkpoint.
Fortuitously, this asymmetry is oriented in the forward direction that is relevant to persistent calibration.
It may be that this asymmetry stems from the tendency for neural networks to be more plastic earlier and more stable later in training \citep{achille2018critical}; this would suggest that features developed at certain points in time tend to remain after that point, but are not present before. An adapter trained on the later checkpoint may rely on those features, which are then absent in the earlier checkpoint, harming its performance, but an adapter trained on the earlier checkpoint can only learn to rely on the features present at that time, which are likely to persist into the later checkpoint. 
In \cref{app:variance-analysis}, we support this by showing that the variance, computed over confidence adapter training seeds, of hidden states for a given QA instance, tends to be higher on later checkpoints, suggesting that later checkpoints have developed a larger space of confidence features for confidence adapters to choose from.

\myparagraph{Random Labels.} We dig deeper into the heterogeneity of this feature space by testing if there exist confidence estimators with maintained performance on a given checkpoint but catastrophic performance on another checkpoint: can we train an adapter to perform on \C2 but \textit{not} on \C3?
If transfer is inevitable, a drop on \C3 should be accompanied by a drop on \C2, i.e.\ one cannot optimize for \C2 without inadvertently optimizing for \C3.
To test this, we run multi-checkpoint training on [\C2, \C3] but with random labels for \C3 -- independent samples from $\rm Bernoulli(0.5)$.
\cref{fig:triviaqa-transfer} (right) reveals that \C3 performance generally plummets, but at no penalty to \C2 performance.
This shows the extreme heterogeneity of the hypothesis space of confidence features that are viable on a given checkpoint, which may hinder identifying generalizable meta-knowledge features for persistent calibration.

\section{Related Work}
\label{sec:related-work}

\myparagraph{Confidence Estimation.} The main families of LLM confidence estimation methods are based on generation consistency \citep{kuhn2023semantic, manakul2023selfcheckgpt}, verbalized confidence \citep{kadavath2022language, wang2026calibrating}, fine-tuning \citep{kapoor2024large, li2026conftuner}, and external verifiers \citep{zhu2026a, xiao2026generalized}.
While trained methods achieve the best calibration in-domain, they often struggle with domain transfer due to domain-specific signals \citep{sky2024androids, srey2026signals, ying2026truthfulness}.
We highlight another dimension of brittleness arising from checkpoint-specific features, which multi-checkpoint training partially alleviates.

\myparagraph{Continual Learning.} As AI systems increasingly act as personalized assistants \citep{uddin2026recall}, scientific collaborators \citep{zhang2025exploring}, and autonomous agents \citep{luo2025large}, they must continually adapt to new developments in human users, the scientific literature, and the world.
The families of continual learning methods for LLMs include model editing, retrieval, and continued pre-training \citep{zheng2026lifelong}.
We focus on pre-training, as it is the phase in which LLMs acquire the vast majority of their knowledge \citep{gekhman2024does, chang2024large}.
\citet{hasegawa2025knowledge} find that model editing leads to underconfidence in token-probability--based confidence.
\citet{liu2026fast} use a frozen probe to monitor task performance during pre-training, showing viable forward transfer.
Our key distinction is that we center our evaluation around knowledge contrast sets to isolate the dependence between knowledge and confidence.
In \cref{app:related-work}, we discuss contrast sets \citep{gardner2020evaluating} and metacognition.

\section{Conclusion}

We motivate \textit{persistent calibration} as a requirement for trustworthy, continually learning models.
To see if confidence estimators can track the evolution of knowledge, we evaluate on \textit{knowledge contrast sets}, which highlight changes in knowledge.
We find substantial headroom, despite the gains from multi-checkpoint training.
Our analysis finds that LLM training dynamics are encouraging for persistent calibration, although the vast space of checkpoint-specific confidence features may hinder identifying robust meta-knowledge features.

\section*{Acknowledgments}
The authors acknowledge support from Coefficient Giving, and acknowledge the Texas Advanced Computing Center (TACC) at The University of Texas at Austin for providing computational resources that have contributed to the research results reported within this paper.

\bibliography{iclr2027_conference}

@article{kasai2023realtime,
  title={Realtime qa: What's the answer right now?},
  author={Kasai, Jungo and Sakaguchi, Keisuke and Le Bras, Ronan and Asai, Akari and Yu, Xinyan and Radev, Dragomir and Smith, Noah A and Choi, Yejin and Inui, Kentaro and others},
  journal={Advances in neural information processing systems},
  volume={36},
  pages={49025--49043},
  year={2023}
}

@inproceedings{zhou2024relying,
  title={Relying on the unreliable: The impact of language models’ reluctance to express uncertainty},
  author={Zhou, Kaitlyn and Hwang, Jena and Ren, Xiang and Sap, Maarten},
  booktitle={Proceedings of the 62nd Annual Meeting of the Association for Computational Linguistics (Volume 1: Long Papers)},
  pages={3623--3643},
  year={2024}
}

@article{steyvers2025large,
  title={What large language models know and what people think they know},
  author={Steyvers, Mark and Tejeda, Heliodoro and Kumar, Aakriti and Belem, Catarina and Karny, Sheer and Hu, Xinyue and Mayer, Lukas W and Smyth, Padhraic},
  journal={Nature Machine Intelligence},
  volume={7},
  number={2},
  pages={221--231},
  year={2025},
  publisher={Nature Publishing Group UK London}
}

@article{liu2026fast,
  title={Fast and Accurate Probing of In-Training LLMs' Downstream Performances},
  author={Liu, Zhichen and Lun, Tianle and Wen, Zhibin and An, Hao and Ou, Yulin and Xu, Jianhui and Zhang, Hao and Fang, Wenyi and Zheng, Yang and Xu, Yang},
  journal={arXiv preprint arXiv:2604.01025},
  year={2026}
}

@inproceedings{yona2026position,
    title={Position: Hallucinations Undermine Trust; Metacognition is a Way Forward},
    author={Gal Yona and Mor Geva and Yossi Matias},
    booktitle={Forty-third International Conference on Machine Learning Position Paper Track},
    year={2026},
    url={https://openreview.net/forum?id=I9CIvYaRmB}
}

@article{steyvers2026metacognition,
  title={Metacognition and uncertainty communication in humans and large language models},
  author={Steyvers, Mark and Peters, Megan AK},
  journal={Current Directions in Psychological Science},
  volume={35},
  number={3},
  pages={131--139},
  year={2026},
  publisher={SAGE Publications Sage CA: Los Angeles, CA}
}

@article{gureckis2012self,
  title={Self-directed learning: A cognitive and computational perspective},
  author={Gureckis, Todd M and Markant, Douglas B},
  journal={Perspectives on Psychological Science},
  volume={7},
  number={5},
  pages={464--481},
  year={2012},
  publisher={Sage Publications Sage CA: Los Angeles, CA}
}

@inproceedings{binder2025looking,
  title={Looking inward: Language models can learn about themselves by introspection},
  author={Binder, Felix Jedidja and Chua, James and Korbak, Tomek and Sleight, Henry and Hughes, John and Long, Robert and Perez, Ethan and Turpin, Miles and Evans, Owain},
  booktitle={International Conference on Learning Representations},
  volume={2025},
  pages={3710--3756},
  year={2025}
}

@article{guo2026introspective,
  title={Introspective Coupling: Self-Explanation Training Tracks Behavioral Change Despite Fixed Supervision},
  author={Guo, Zifan Carl and Ruis, Laura and Andreas, Jacob and Li, Belinda Z},
  journal={arXiv preprint arXiv:2606.32038},
  year={2026}
}

@article{zhang2025exploring,
  title={Exploring the role of large language models in the scientific method: from hypothesis to discovery},
  author={Zhang, Yanbo and Khan, Sumeer A and Mahmud, Adnan and Yang, Huck and Lavin, Alexander and Levin, Michael and Frey, Jeremy and Dunnmon, Jared and Evans, James and Bundy, Alan and others},
  journal={npj Artificial Intelligence},
  volume={1},
  number={1},
  pages={14},
  year={2025},
  publisher={Nature Publishing Group UK London}
}

@book{ring1994continual,
  title={Continual learning in reinforcement environments},
  author={Ring, Mark Bishop},
  year={1994},
  publisher={The University of Texas at Austin}
}

@article{shi2025continual,
  title={Continual learning of large language models: A comprehensive survey},
  author={Shi, Haizhou and Xu, Zihao and Wang, Hengyi and Qin, Weiyi and Wang, Wenyuan and Wang, Yibin and Wang, Zifeng and Ebrahimi, Sayna and Wang, Hao},
  journal={ACM Computing Surveys},
  volume={58},
  number={5},
  pages={1--42},
  year={2025},
  publisher={ACM New York, NY}
}

@article{bai2022constitutional,
  title={Constitutional ai: Harmlessness from ai feedback},
  author={Bai, Yuntao and Kadavath, Saurav and Kundu, Sandipan and Askell, Amanda and Kernion, Jackson and Jones, Andy and Chen, Anna and Goldie, Anna and Mirhoseini, Azalia and McKinnon, Cameron and others},
  journal={arXiv preprint arXiv:2212.08073},
  year={2022}
}

@article{wynn2024learning,
  title={Learning human-like representations to enable learning human values},
  author={Wynn, Andrea H and Sucholutsky, Ilia and Griffiths, Thomas L},
  journal={Advances in Neural Information Processing Systems},
  volume={37},
  pages={30230--30260},
  year={2024}
}

@article{wang2023trace,
  title={Trace: A comprehensive benchmark for continual learning in large language models},
  author={Wang, Xiao and Zhang, Yuansen and Chen, Tianze and Gao, Songyang and Jin, Senjie and Yang, Xianjun and Xi, Zhiheng and Zheng, Rui and Zou, Yicheng and Gui, Tao and others},
  journal={arXiv preprint arXiv:2310.06762},
  year={2023}
}

@article{kapoor2024large,
  title={Large language models must be taught to know what they don’t know},
  author={Kapoor, Sanyam and Gruver, Nate and Roberts, Manley and Collins, Katherine and Pal, Arka and Bhatt, Umang and Weller, Adrian and Dooley, Samuel and Goldblum, Micah and Wilson, Andrew G},
  journal={Advances in Neural Information Processing Systems},
  volume={37},
  pages={85932--85972},
  year={2024}
}

@article{li2026conftuner,
  title={Conftuner: Training large language models to express their confidence verbally},
  author={Li, Yibo and Xiong, Miao and Wu, Jiaying and Hooi, Bryan},
  journal={Advances in Neural Information Processing Systems},
  volume={38},
  pages={53484--53513},
  year={2026}
}

@article{srey2026signals,
  title={From Signals to Transfer: A Factorised Study of Probe-Based Uncertainty Estimation in Large Language Models},
  author={Srey, Ponhvoan and Wu, Xiaobao and Nguyen, Cong-Duy and Nguyen, Quang Minh and Vu, Duc Anh and Luu, Anh Tuan},
  journal={arXiv preprint arXiv:2606.27679},
  year={2026}
}

@article{ying2026truthfulness,
  title={The truthfulness spectrum hypothesis},
  author={Ying, Zhuofan Josh and Ravfogel, Shauli and Kriegeskorte, Nikolaus and Hase, Peter},
  journal={arXiv preprint arXiv:2602.20273},
  year={2026}
}

@inproceedings{kuhn2023semantic,
    title={Semantic Uncertainty: Linguistic Invariances for Uncertainty Estimation in Natural Language Generation},
    author={Lorenz Kuhn and Yarin Gal and Sebastian Farquhar},
    booktitle={The Eleventh International Conference on Learning Representations },
    year={2023},
    url={https://openreview.net/forum?id=VD-AYtP0dve}
}

@inproceedings{manakul2023selfcheckgpt,
  title={Selfcheckgpt: Zero-resource black-box hallucination detection for generative large language models},
  author={Manakul, Potsawee and Liusie, Adian and Gales, Mark},
  booktitle={Proceedings of the 2023 conference on empirical methods in natural language processing},
  pages={9004--9017},
  year={2023}
}

@inproceedings{wang2026calibrating,
  title={Calibrating verbalized confidence with self-generated distractors},
  author={Wang, Victor and Stengel-Eskin, Elias},
  booktitle={International Conference on Learning Representations},
  volume={2026},
  pages={121268--121297},
  year={2026}
}

@article{chow1957optimum,
  title={An optimum character recognition system using decision functions},
  author={Chow, Chi-Keung},
  journal={IRE Transactions on Electronic Computers},
  number={4},
  pages={247--254},
  year={1957},
  publisher={IEEE}
}

@inproceedings{wang2026position,
    title={Position: Agents Should Invoke External Tools {ONLY} When Epistemically Necessary},
    author={Hongru Wang and Cheng Qian and Manling Li and Jiahao Qiu and Boyang XUE and Mengdi Wang and Heng Ji and Amos Storkey and Kam-Fai Wong},
    booktitle={Forty-third International Conference on Machine Learning Position Paper Track},
    year={2026},
    url={https://openreview.net/forum?id=5wTHg2zQRA}
}

@inproceedings{asai2024self,
  title={Self-rag: Learning to retrieve, generate, and critique through self-reflection},
  author={Asai, Akari and Wu, Zeqiu and Wang, Yizhong and Sil, Avi and Hajishirzi, Hannaneh},
  booktitle={International conference on learning representations},
  volume={2024},
  pages={9112--9141},
  year={2024}
}

@article{guo2025deepseek,
  title={Deepseek-r1: Incentivizing reasoning capability in llms via reinforcement learning},
  author={Guo, Daya and Yang, Dejian and Zhang, Haowei and Song, Junxiao and Wang, Peiyi and Zhu, Qihao and Xu, Runxin and Zhang, Ruoyu and Ma, Shirong and Bi, Xiao and others},
  journal={arXiv preprint arXiv:2501.12948},
  year={2025}
}

@inproceedings{gururangan2020don,
  title={Don’t stop pretraining: Adapt language models to domains and tasks},
  author={Gururangan, Suchin and Marasovi{\'c}, Ana and Swayamdipta, Swabha and Lo, Kyle and Beltagy, Iz and Downey, Doug and Smith, Noah A},
  booktitle={Proceedings of the 58th annual meeting of the association for computational linguistics},
  pages={8342--8360},
  year={2020}
}

@article{wang2025parameter,
  title={Parameter-efficient fine-tuning in large language models: a survey of methodologies},
  author={Wang, Luping and Chen, Sheng and Jiang, Linnan and Pan, Shu and Cai, Runze and Yang, Sen and Yang, Fei},
  journal={Artificial Intelligence Review},
  volume={58},
  number={8},
  pages={227},
  year={2025},
  publisher={Springer}
}

@inproceedings{zhu2023calibration,
  title={On the calibration of large language models and alignment},
  author={Zhu, Chiwei and Xu, Benfeng and Wang, Quan and Zhang, Yongdong and Mao, Zhendong},
  booktitle={Findings of the Association for Computational Linguistics: EMNLP 2023},
  pages={9778--9795},
  year={2023}
}

@article{luo2026your,
  title={Your pre-trained LLM is secretly an unsupervised confidence calibrator},
  author={Luo, Beier and Wang, Shuoyuan and Li, Sharon and Wei, Hongxin},
  journal={Advances in Neural Information Processing Systems},
  volume={38},
  pages={163514--163543},
  year={2026}
}

@article{zhu2026a,
    title={A Systematic Assessment of Weak-to-Strong Confidence Prediction in Large Language Models},
    author={Tracy Yixin Zhu and Yukai Yang and Marco Morucci and Tim G. J. Rudner},
    journal={Transactions on Machine Learning Research},
    issn={2835-8856},
    year={2026},
    url={https://openreview.net/forum?id=xYSzkg5qPD},
}

@inproceedings{xiao2026generalized,
    title={Generalized Correctness Models: Learning Calibrated and Cross-Model Correctness Predictors from Historical Patterns},
    author={Hanqi Xiao and Vaidehi Patil and Hyunji Lee and Elias Stengel-Eskin and Mohit Bansal},
    booktitle={Forty-third International Conference on Machine Learning},
    year={2026},
    url={https://openreview.net/forum?id=g9G7qyAzki}
}

@misc{periodiclabs2026nature,
  author       = {{Periodic Labs}},
  title        = {Nature Is Our Learning Environment},
  year         = {2026},
  month        = sep,
  day          = {15},
  howpublished = {\url{https://periodic.com/news/nature-is-our-learning-environment}},
  note         = {Accessed: 2026-09-15}
}

@inproceedings{mallen2023not,
  title={When not to trust language models: Investigating effectiveness of parametric and non-parametric memories},
  author={Mallen, Alex and Asai, Akari and Zhong, Victor and Das, Rajarshi and Khashabi, Daniel and Hajishirzi, Hannaneh},
  booktitle={Proceedings of the 61st annual meeting of the association for computational linguistics (volume 1: Long papers)},
  pages={9802--9822},
  year={2023}
}

@article{kadavath2022language,
  title={Language models (mostly) know what they know},
  author={Kadavath, Saurav and Conerly, Tom and Askell, Amanda and Henighan, Tom and Drain, Dawn and Perez, Ethan and Schiefer, Nicholas and Hatfield-Dodds, Zac and DasSarma, Nova and Tran-Johnson, Eli and others},
  journal={arXiv preprint arXiv:2207.05221},
  year={2022}
}

@inproceedings{tian2023just,
  title={Just ask for calibration: Strategies for eliciting calibrated confidence scores from language models fine-tuned with human feedback},
  author={Tian, Katherine and Mitchell, Eric and Zhou, Allan and Sharma, Archit and Rafailov, Rafael and Yao, Huaxiu and Finn, Chelsea and Manning, Christopher D},
  booktitle={Proceedings of the 2023 Conference on Empirical Methods in Natural Language Processing},
  pages={5433--5442},
  year={2023}
}

@inproceedings{sky2024androids,
  title={Do androids know they’re only dreaming of electric sheep?},
  author={Sky, CH-Wang and Van Durme, Benjamin and Eisner, Jason and Kedzie, Chris},
  booktitle={Findings of the Association for Computational Linguistics: ACL 2024},
  pages={4401--4420},
  year={2024}
}

@article{geirhos2020shortcut,
  title={Shortcut learning in deep neural networks},
  author={Geirhos, Robert and Jacobsen, J{\"o}rn-Henrik and Michaelis, Claudio and Zemel, Richard and Brendel, Wieland and Bethge, Matthias and Wichmann, Felix A},
  journal={Nature Machine Intelligence},
  volume={2},
  number={11},
  pages={665--673},
  year={2020},
  publisher={Nature Publishing Group UK London}
}

@article{d2022underspecification,
  title={Underspecification presents challenges for credibility in modern machine learning},
  author={D'Amour, Alexander and Heller, Katherine and Moldovan, Dan and Adlam, Ben and Alipanahi, Babak and Beutel, Alex and Chen, Christina and Deaton, Jonathan and Eisenstein, Jacob and Hoffman, Matthew D and others},
  journal={Journal of Machine Learning Research},
  volume={23},
  number={226},
  pages={1--61},
  year={2022}
}

@inproceedings{gardner2020evaluating,
  title={Evaluating models’ local decision boundaries via contrast sets},
  author={Gardner, Matt and Artzi, Yoav and Basmov, Victoria and Berant, Jonathan and Bogin, Ben and Chen, Sihao and Dasigi, Pradeep and Dua, Dheeru and Elazar, Yanai and Gottumukkala, Ananth and others},
  booktitle={Findings of the Association for Computational Linguistics: EMNLP 2020},
  pages={1307--1323},
  year={2020}
}

@inproceedings{li2020linguistically,
  title={Linguistically-informed transformations (LIT): A method for automatically generating contrast sets},
  author={Li, Chuanrong and Shengshuo, Lin and Liu, Zeyu and Wu, Xinyi and Zhou, Xuhui and Steinert-Threlkeld, Shane},
  booktitle={Proceedings of the third BlackboxNLP workshop on analyzing and interpreting neural networks for NLP},
  pages={126--135},
  year={2020}
}

@inproceedings{bitton2021automatic,
  title={Automatic generation of contrast sets from scene graphs: Probing the compositional consistency of GQA},
  author={Bitton, Yonatan and Stanovsky, Gabriel and Schwartz, Roy and Elhadad, Michael},
  booktitle={Proceedings of the 2021 conference of the North American chapter of the association for computational linguistics: human language technologies},
  pages={94--105},
  year={2021}
}

@inproceedings{shekhar2017foil,
  title={Foil it! find one mismatch between image and language caption},
  author={Shekhar, Ravi and Pezzelle, Sandro and Klimovich, Yauhen and Herbelot, Aur{\'e}lie and Nabi, Moin and Sangineto, Enver and Bernardi, Raffaella},
  booktitle={Proceedings of the 55th Annual Meeting of the Association for Computational Linguistics (Volume 1: Long Papers)},
  pages={255--265},
  year={2017}
}

@inproceedings{marvin2018targeted,
  title={Targeted syntactic evaluation of language models},
  author={Marvin, Rebecca and Linzen, Tal},
  booktitle={Proceedings of the 2018 conference on empirical methods in natural language processing},
  pages={1192--1202},
  year={2018}
}

@inproceedings{rudinger2018gender,
  title={Gender bias in coreference resolution},
  author={Rudinger, Rachel and Naradowsky, Jason and Leonard, Brian and Van Durme, Benjamin},
  booktitle={Proceedings of the 2018 Conference of the North American Chapter of the Association for Computational Linguistics: Human Language Technologies, Volume 2 (Short Papers)},
  pages={8--14},
  year={2018}
}

@article{sakaguchi2021winogrande,
  title={Winogrande: An adversarial winograd schema challenge at scale},
  author={Sakaguchi, Keisuke and Bras, Ronan Le and Bhagavatula, Chandra and Choi, Yejin},
  journal={Communications of the ACM},
  volume={64},
  number={9},
  pages={99--106},
  year={2021},
  publisher={ACM New York, NY, USA}
}

@InProceedings{huang2025math,
  title = 	 {{MATH}-Perturb: Benchmarking {LLM}s’ Math Reasoning Abilities against Hard Perturbations},
  author =       {Huang, Kaixuan and Guo, Jiacheng and Li, Zihao and Ji, Xiang and Ge, Jiawei and Li, Wenzhe and Guo, Yingqing and Cai, Tianle and Yuan, Hui and Wang, Runzhe and Wu, Yue and Yin, Ming and Tang, Shange and Huang, Yangsibo and Jin, Chi and Chen, Xinyun and Zhang, Chiyuan and Wang, Mengdi},
  booktitle = 	 {Proceedings of the 42nd International Conference on Machine Learning},
  pages = 	 {25311--25328},
  year = 	 {2025},
  editor = 	 {Singh, Aarti and Fazel, Maryam and Hsu, Daniel and Lacoste-Julien, Simon and Berkenkamp, Felix and Maharaj, Tegan and Wagstaff, Kiri and Zhu, Jerry},
  volume = 	 {267},
  series = 	 {Proceedings of Machine Learning Research},
  month = 	 {13--19 Jul},
  publisher =    {PMLR},
  url = 	 {https://proceedings.mlr.press/v267/huang25k.html},
}

@inproceedings{mirzadeh2025gsm,
  title={Gsm-symbolic: Understanding the limitations of mathematical reasoning in large language models},
  author={Mirzadeh, Iman and Alizadeh-Vahid, Keivan and Shahrokhi, Hooman and Tuzel, Oncel and Bengio, Samy and Farajtabar, Mehrdad},
  booktitle={International Conference on Learning Representations},
  volume={2025},
  pages={94743--94765},
  year={2025}
}

@inproceedings{ashuach2026masked,
  title={Masked by Consensus: Disentangling Privileged Knowledge in LLM Correctness},
  author={Ashuach, Tomer and Gretz, Shai and Katz, Yoav and Belinkov, Yonatan and Ein-Dor, Liat},
  booktitle={Proceedings of the 64th Annual Meeting of the Association for Computational Linguistics (Volume 1: Long Papers)},
  pages={10577--10596},
  year={2026}
}

@inproceedings{uddin2026recall,
  title={From recall to forgetting: Benchmarking long-term memory for personalized agents},
  author={Uddin, Md Nayem and Shubham, Kumar and Blanco, Eduardo and Baral, Chitta and Wang, Gengyu},
  booktitle={Findings of the Association for Computational Linguistics: ACL 2026},
  pages={26814--26841},
  year={2026}
}

@article{luo2025large,
  title={Large language model agent: A survey on methodology, applications and challenges},
  author={Luo, Junyu and Zhang, Weizhi and Yuan, Ye and Zhao, Yusheng and Yang, Junwei and Gu, Yiyang and Wu, Bohan and Chen, Binqi and Qiao, Ziyue and Long, Qingqing and others},
  journal={arXiv preprint arXiv:2503.21460},
  year={2025}
}

@article{zheng2026lifelong,
  title={Lifelong learning of large language model based agents: A roadmap},
  author={Zheng, Junhao and Shi, Chengming and Cai, Xidi and Li, Qiuke and Zhang, Duzhen and Li, Chenxing and Yu, Dong and Ma, Qianli},
  journal={IEEE Transactions on Pattern Analysis and Machine Intelligence},
  year={2026},
  publisher={IEEE}
}

@article{chang2024large,
  title={How do large language models acquire factual knowledge during pretraining?},
  author={Chang, Hoyeon and Park, Jinho and Ye, Seonghyeon and Yang, Sohee and Seo, Youngkyung and Chang, Du-Seong and Seo, Minjoon},
  journal={Advances in neural information processing systems},
  volume={37},
  pages={60626--60668},
  year={2024}
}

@inproceedings{gekhman2024does,
  title={Does fine-tuning LLMs on new knowledge encourage hallucinations?},
  author={Gekhman, Zorik and Yona, Gal and Aharoni, Roee and Eyal, Matan and Feder, Amir and Reichart, Roi and Herzig, Jonathan},
  booktitle={Proceedings of the 2024 conference on empirical methods in natural language processing},
  pages={7765--7784},
  year={2024}
}

@article{olmo2025olmo,
  title={Olmo 3},
  author={Olmo, Team and Ettinger, Allyson and Bertsch, Amanda and Kuehl, Bailey and Graham, David and Heineman, David and Groeneveld, Dirk and Brahman, Faeze and Timbers, Finbarr and Ivison, Hamish and others},
  journal={arXiv preprint arXiv:2512.13961},
  year={2025}
}

@misc{marin,
  author       = {{Marin Community}},
  title        = {Marin: Open-source framework for the research and development of foundation models},
  year         = {2026},
  howpublished = {\url{https://github.com/marin-community/marin}},
}

@inproceedings{joshi2017triviaqa,
  title={Triviaqa: A large scale distantly supervised challenge dataset for reading comprehension},
  author={Joshi, Mandar and Choi, Eunsol and Weld, Daniel S and Zettlemoyer, Luke},
  booktitle={Proceedings of the 55th Annual Meeting of the Association for Computational Linguistics (Volume 1: Long Papers)},
  pages={1601--1611},
  year={2017}
}

@misc{openaccessaicollective_jeopardy,
  author       = {{Open Access AI Collective}},
  title        = {Jeopardy},
  howpublished = {Hugging Face Datasets},
  url          = {https://huggingface.co/datasets/openaccess-ai-collective/jeopardy},
  note         = {Accessed: 2026-09-16},
  year         = {2023},
}

@article{krithara2023bioasq,
  title={BioASQ-QA: A manually curated corpus for Biomedical Question Answering},
  author={Krithara, Anastasia and Nentidis, Anastasios and Bougiatiotis, Konstantinos and Paliouras, Georgios},
  journal={Scientific data},
  volume={10},
  number={1},
  pages={170},
  year={2023},
  publisher={Nature Publishing Group UK London}
}

@inproceedings{hasegawa2025knowledge,
  title={Knowledge Editing Induces Underconfidence in Language Models},
  author={Hasegawa, Ryo and Sakai, Yusuke and Kamigaito, Hidetaka and Watanabe, Taro},
  booktitle={Proceedings of the 14th Joint Conference on Lexical and Computational Semantics (* SEM 2025)},
  pages={338--347},
  year={2025}
}

@article{hanley1982meaning,
  title={The meaning and use of the area under a receiver operating characteristic (ROC) curve.},
  author={Hanley, James A and McNeil, Barbara J},
  journal={Radiology},
  volume={143},
  number={1},
  pages={29--36},
  year={1982}
}

@article{brier1950verification,
  title={Verification of forecasts expressed in terms of probability},
  author={Glenn W. Brier},
  journal={Monthly Weather Review},
  year={1950},
  volume={78},
  pages={1-3},
  url={https://api.semanticscholar.org/CorpusID:122906757}
}

@inproceedings{naeini2015obtaining,
  title={Obtaining well calibrated probabilities using bayesian binning},
  author={Naeini, Mahdi Pakdaman and Cooper, Gregory and Hauskrecht, Milos},
  booktitle={Proceedings of the AAAI conference on artificial intelligence},
  volume={29},
  year={2015}
}

@inproceedings{blasiok2024smooth,
  title={Smooth ECE: Principled reliability diagrams via kernel smoothing},
  author={Blasiok, Jaroslaw and Nakkiran, Preetum},
  booktitle={International Conference on Learning Representations},
  volume={2024},
  pages={1679--1707},
  year={2024}
}

@article{singh2026can,
  title={Can LLMs Introspect? A Reality Check},
  author={Singh, Shashwat and Linzen, Tal and Ravfogel, Shauli},
  journal={arXiv preprint arXiv:2605.26242},
  year={2026}
}

@article{nixon2019measuring,
  title={Measuring calibration in deep learning},
  author={Nixon, Jeremy and Dusenberry, Mike and Jerfel, Ghassen and Nguyen, Timothy and Liu, Jeremiah and Zhang, Linchuan and Tran, Dustin},
  journal={arXiv preprint arXiv:1904.01685},
  year={2019}
}

@inproceedings{fadeeva2026don,
  title={Don't Throw Away Your Beams: Improving Consistency-based Uncertainties in LLMs via Beam Search},
  author={Fadeeva, Ekaterina and Goloburda, Maiya and Rubashevskii, Aleksandr and Vashurin, Roman and Shelmanov, Artem and Nakov, Preslav and Sachan, Mrinmaya and Panov, Maxim},
  booktitle={International Conference on Learning Representations},
  volume={2026},
  pages={139866--139895},
  year={2026}
}

@inproceedings{hu2022lora,
    title={Lo{RA}: Low-Rank Adaptation of Large Language Models},
    author={Edward J Hu and yelong shen and Phillip Wallis and Zeyuan Allen-Zhu and Yuanzhi Li and Shean Wang and Lu Wang and Weizhu Chen},
    booktitle={International Conference on Learning Representations},
    year={2022},
    url={https://openreview.net/forum?id=nZeVKeeFYf9}
}

@article{wei2024measuring,
  title={Measuring short-form factuality in large language models},
  author={Wei, Jason and Karina, Nguyen and Chung, Hyung Won and Jiao, Yunxin Joy and Papay, Spencer and Glaese, Amelia and Schulman, John and Fedus, William},
  journal={arXiv preprint arXiv:2411.04368},
  year={2024}
}

@inproceedings{achille2018critical,
  title={Critical learning periods in deep networks},
  author={Achille, Alessandro and Rovere, Matteo and Soatto, Stefano},
  booktitle={International conference on learning representations},
  year={2018}
}

@article{bradley1952rank,
  title={Rank analysis of incomplete block designs: I. the method of paired comparisons},
  author={Bradley, Ralph Allan and Terry, Milton E},
  journal={Biometrika},
  volume={39},
  number={3/4},
  pages={324--345},
  year={1952},
  publisher={JSTOR}
}

@article{sun2024rethinking,
  title={Rethinking bradley-terry models in preference-based reward modeling: Foundations, theory, and alternatives},
  author={Sun, Hao and Shen, Yunyi and Ton, Jean-Francois},
  journal={arXiv preprint arXiv:2411.04991},
  year={2024}
}

@article{christiano2017deep,
  title={Deep reinforcement learning from human preferences},
  author={Christiano, Paul F and Leike, Jan and Brown, Tom and Martic, Miljan and Legg, Shane and Amodei, Dario},
  journal={Advances in neural information processing systems},
  volume={30},
  year={2017}
}

@article{ouyang2022training,
  title={Training language models to follow instructions with human feedback},
  author={Ouyang, Long and Wu, Jeffrey and Jiang, Xu and Almeida, Diogo and Wainwright, Carroll and Mishkin, Pamela and Zhang, Chong and Agarwal, Sandhini and Slama, Katarina and Ray, Alex and others},
  journal={Advances in neural information processing systems},
  volume={35},
  pages={27730--27744},
  year={2022}
}

@article{rafailov2023direct,
  title={Direct preference optimization: Your language model is secretly a reward model},
  author={Rafailov, Rafael and Sharma, Archit and Mitchell, Eric and Manning, Christopher D and Ermon, Stefano and Finn, Chelsea},
  journal={Advances in neural information processing systems},
  volume={36},
  pages={53728--53741},
  year={2023}
}

@article{loshchilov2017decoupled,
  title={Decoupled weight decay regularization},
  author={Loshchilov, Ilya and Hutter, Frank},
  journal={arXiv preprint arXiv:1711.05101},
  year={2017}
}

@inproceedings{huang2017snapshot,
    title={Snapshot Ensembles: Train 1, Get M for Free},
    author={Gao Huang and Yixuan Li and Geoff Pleiss and Zhuang Liu and John E. Hopcroft and Kilian Q. Weinberger},
    booktitle={International Conference on Learning Representations},
    year={2017},
    url={https://openreview.net/forum?id=BJYwwY9ll}
}

@article{garipov2018loss,
  title={Loss surfaces, mode connectivity, and fast ensembling of dnns},
  author={Garipov, Timur and Izmailov, Pavel and Podoprikhin, Dmitrii and Vetrov, Dmitry and Wilson, Andrew G},
  journal={Advances in neural information processing systems},
  volume={31},
  year={2018}
}
\bibliographystyle{iclr2027_conference}

\appendix

\section{Experimental Setup}
\label{app:experimental-setup}

\subsection{Conditioning on Contrast Set Membership}
\label{app:condition-conset}

Since a contrast set derived from a full dataset is not a random subset but rather based on correctness labels, confidence estimates that are calibrated with respect to the full data distribution may not be calibrated with respect to the distribution that conditions on contrast set membership.
For example, suppose that on each question in a dataset, two checkpoints each have an independent 80\% chance of predicting the correct answer.
The ideal confidence estimates are thus 80\% on every predicted answer, and the expected accuracy is 80\%.
The contrast set consists of questions for which either checkpoint 1 is correct and checkpoint 2 is incorrect ($0.8 \cdot 0.2 = 0.16$ of all instances), or checkpoint 1 is incorrect and checkpoint 2 is correct ($0.2 \cdot 0.8 = 0.16$ of all instances), so each checkpoint has an expected accuracy of 50\% on contrast set questions.
Thus, the ideal global confidence estimates of 80\% appear miscalibrated when viewed within the contrast set.

To account for the selection bias of the contrast set, it is necessary to normalize the confidence estimates for the two answers to a given question in the contrast set, as illustrated in \cref{fig:main} (right).
This normalization should restore contrast-set calibration for confidence estimates that are calibrated with respect to the full dataset; accordingly, we treat confidence estimates as if they were calibrated.
In the example above, every (0.8, 0.8) confidence pair should be normalized to (0.5, 0.5) because each answer has a 50\% chance of being correct given that exactly one of the two answers is correct.
Let checkpoint $k$'s answer have binary correctness $z_k$ and calibrated confidence $p_k = P(z_k = 1)$.
We would like to find
\begin{align*}
    P(z_1 = 1 \mid \underbrace{(z_1, z_2) \in \{(1, 0), (0, 1)\}}_{\text{contrast set membership}}) = \frac{P((z_1, z_2) = (1, 0))}{P((z_1, z_2) \in \{(1, 0), (0, 1)\})} .
\end{align*}
If we assume the checkpoints' predictions are independent as in the example above, this becomes
\begin{align}
    \frac{P((z_1, z_2) = (1, 0))}{P((z_1, z_2) \in \{(1, 0), (0, 1)\})} = \frac{p_1(1 - p_2)}{p_1(1 - p_2) + (1 - p_1)p_2} . \label{eq:condition-conset-independent}
\end{align}
In general, two checkpoints' predictions are not independent because the later checkpoint originates from the earlier checkpoint, which means that predictions carry over by default.
For example, if checkpoint 1 has 80\% confidence in its answer to a question, and checkpoint 2 is obtained by further training on a negligible amount of unrelated data, it will have a nearly unchanged 80\% confidence on the same answer.
In this case, we should estimate the confidence that both answers are correct as 0.8 rather than the $0.8^2 = 0.64$ that independence would prescribe.
Nonetheless, this ``carry-over'' dependence is not present when the two checkpoints produce \textit{different} predictions, which is the case for contrast set questions by construction.
Thus, we use the normalization given by \cref{eq:condition-conset-independent}, making the assumption that the relative probabilities of the contrast outcomes $(z_1, z_2) = (1, 0), (0, 1)$ are the same as if $z_1$ and $z_2$ were independent.
We provide empirical support for this decision in \cref{app:learn-condition-conset} by showing that \cref{eq:condition-conset-independent} and a fitted normalization function produce similar results.
Moreover, the metrics $\Delta_0, \Delta^b_0$ (\cref{sec:persistent-calibration}) simply compare whether $p_1 > p_2$ (unaffected by normalization), and our results show that they exhibit the same trends as the other metrics, which do depend on normalization.

\subsection{Evaluation Metrics}
\label{app:eval-metrics}

Discriminability refers to how well confidence estimates separate correct answers from incorrect ones, which is useful for selective prediction, i.e.\ abstaining on uncertain questions while maintaining high recall and acc uracy of questions answered \citep{chow1957optimum}.
Calibration refers to whether, for all $p$, the accuracy over answers with confidence $p$ is $p$ \citep{naeini2015obtaining}.

AUC measures discriminability, as it can be characterized as the proportion of (correct, incorrect) pairs in which the correct answer is assigned a higher confidence, with ties given half credit, so trivial performance is 0.5 \citep{hanley1982meaning}.

BS is the mean squared error between confidence estimates and the binary correctnesses \citep{brier1950verification}.
Since BS is a proper scoring rule, i.e.\ its expected value is minimized when confidence estimates match the ground truth probabilities of correctness, it measures calibration as well as discriminability (since it rewards a confidence estimate for being close to the correctness, per instance).

ECE measures calibration \citep{naeini2015obtaining}: $\mathbb{E}_{(x, y, \hat{y}) \sim \mathcal{D},\ p = f(x, \hat{y})} |P(\hat{y} = y \mid p) - p|$, where $\mathcal{D}$ is a data distribution of (question $x$, reference answer $y$, predicted answer $\hat{y}$) instances and $f$ is a confidence estimator.
To empirically estimate ECE, it is necessary to bin or smooth the data; we use SmoothECE for its continuous and hyperparameter-free properties \citep{blasiok2024smooth}.

\subsection{Aggregation of Metrics over Checkpoints}
\label{app:agg-over-ckpts}

Since we evaluate on multiple future checkpoints for a robust evaluation, we must specify how we aggregate over them when reporting calibration metrics.

For full-set evaluation, we compute AUC and BS on the pooled data from all checkpoints, while we average the ECE computed on each checkpoint's data.
We do not pool data for ECE to avoid cancellation effects that hide miscalibration \citep{nixon2019measuring}; for example, if checkpoints 1 and 2 have 50\% and 70\% accuracy, respectively, a confidence estimator that outputs a constant 60\% would be perfectly calibrated on the pooled data (ECE = 0), but evaluating on each checkpoint separately would expose the miscalibration (ECE = 0.1).\footnote{While, in general, we cannot eliminate other sources of cancellation that occur even within a single checkpoint, we consider it prudent to eliminate the addressable case of cross-checkpoint cancellation.}

For contrast-set evaluation, we average a given metric computed on the contrast set of each pair of evaluation checkpoints.
We do not pool data over different checkpoint pairs' contrast sets because the normalization (\cref{app:condition-conset}) assumes that a given answer \textit{pair} contains exactly one correct answer.
For a given contrast set, we compute AUC and BS on the (normalized) confidences from the two checkpoints, while we compute ECE on only one checkpoint's confidences to avoid cancellation effects.\footnote{For example, a constant confidence estimator would yield [0.5, 0.5] normalized confidences on every contrast-set question. Since contrast-set accuracy (pooled over the two checkpoints) is always 50\%, we would get ECE = 0, even if the two checkpoints' contrast-set accuracies are 40\% and 60\%, whereas single-checkpoint ECE would expose ECE = 0.1. Note that single-checkpoint ECE is invariant to which of the two checkpoints is used, since for a given question, both the confidences and the correctnesses sum to 1.}

To require sufficient sample size, we only use a contrast set if it contains at least some minimum number of questions.
We set this threshold to 500 for TriviaQA and Jeopardy, and 300 for BioASQ due to its smaller size; see \cref{app:conset-stats} for data statistics.

\subsection{Data Processing}
\label{app:data-processing}

\begin{table}[t]
\caption{Jeopardy categories included in our evaluation.}
\label{tab:jeopardy-categories}
\begin{center}
\small
\begin{tabular}{lll}
\toprule
19th Century America & 3-Letter Words & 4-Letter Words \\
American History & American Literature & Americana \\
Animals & Annual Events & Architecture \\
Around The World & Art & Art \& Artists \\
Astronomy & Authors & Awards \\
Ballet & Biology & Bodies Of Water \\
Books \& Authors & Business \& Industry & Classical Music \\
Colleges \& Universities & Composers & European History \\
Explorers & Famous Americans & Fashion \\
Fictional Characters & First Ladies & Food \\
Food \& Drink & Fruits \& Vegetables & Geography \\
Historic Names & History & Hodgepodge \\
Holidays \& Observances & In The Dictionary & Islands \\
Languages & Literature & Magazines \\
Medicine & Mountains & Museums \\
Music & Musical Instruments & Mythology \\
Nature & Nonfiction & Opera \\
Organizations & People & Poets \& Poetry \\
Pop Music & Potent Potables & Potpourri \\
Quotations & Religion & Science \\
Science \& Nature & Scientists & Shakespeare \\
Sports & State Capitals & Television \\
The Bible & The Body Human & The Civil War \\
The Movies & Theatre & Transportation \\
Travel \& Tourism & U.S. Cities & U.S. Geography \\
U.S. History & U.S. Presidents & Vocabulary \\
Weights \& Measures & Word Origins & World Capitals \\
World Cities & World Geography & World History \\
Zoology &  &  \\
\bottomrule
\end{tabular}
\end{center}
\end{table}

\begin{table*}[t]
\caption{BioASQ example questions. For list questions, an answer matching any of the reference answers is accepted.}
\label{tab:bioasq-examples}
\begin{center}
\small
\begin{tabular}{lp{0.57\textwidth}p{0.25\textwidth}}
\toprule
Type & Question & Reference answer(s) \\
\midrule
Factoid & What is the main use of ETD fragmentation? & Analysis of intact proteins \\
Factoid & Which disease can be treated with Delamanid? & tuberculosis \\
List & List approved radioprotective compounds. Just output one of the answers. & [[`amifostine'], [`palifermin']] \\
List & List the 5 different human immunoglobulin heavy chains. Just output one of the answers. & [[`alpha'], [`delta'], [`epsilon'], [`gamma'], [`mu']] \\
\bottomrule
\end{tabular}
\end{center}
\end{table*}

We use the datasets TriviaQA \citep{joshi2017triviaqa}, Jeopardy \citep{openaccessaicollective_jeopardy}, and BioASQ \citep{krithara2023bioasq}.
Here, we describe how we process the data before use.

\myparagraph{TriviaQA and Jeopardy.} For each dataset, we shuffle the questions and remove duplicate questions.
We use the validation split of the \texttt{rc.nocontext} subset of TriviaQA.
The Jeopardy dataset contains many categories, not all of which are suitable for factual recall.
We exclude questions containing HTML media links.
We manually inspect the categories with at least 150 questions and check for the suitability of each category.
We exclude the following categories: Before \& After, Rhyme Time, Stupid Answers, Common Bonds, Homophones.
See \cref{tab:jeopardy-categories} for the list of included categories.
We prepend the question category to the question.
After deduplication, for each of TriviaQA and Jeopardy, we reserve 10 questions for potential ICL use (excluding them from training and evaluation data), although we use 5 ICL examples in practice.
TriviaQA is left with 9961 questions, which we split into a train set of 5000 questions and a test set of 4961 questions.
Jeopardy has a train set of 5000 questions and a test set of 5000 questions.
In a manual audit, we identified several low-quality questions, which we blacklisted.
After blacklisting, TriviaQA's train and test splits have sizes 4999 and 4960, respectively, and Jeopardy's train and test splits have sizes 4997 and 4998, respectively.

\myparagraph{BioASQ.} We use the BioASQ 14b training data, retaining factoid and list questions only, as these are suitable for factual recall evaluation.
Factoid questions have a short-form answer, and list questions request a list of multiple answers; see \cref{tab:bioasq-examples} for examples.
We inspected reference answers for several list questions and found that they often do not produce an exhaustive list, so we only require models to provide any correct answer, by appending \textit{``Just output one of the answers.''} to the question.
After shuffling and deduplicating questions, we discard list questions with more than 10 answer groups in order to not overwhelm the LLM judge, which is tasked with checking if the candidate answer matches any of the reference answers.
For each question type (factoid or list), we reserve 5 questions as ICL examples.
For the list-type ICL examples, we set the example answer to the first answer in the reference list, as we find that this is generally the most noteworthy answer.
For evaluation, we are left with 1690 factoid questions and 1029 list questions, for a total of 2719 questions.

\subsection{Audit of Data Quality}
\label{app:audit-data-quality}

We use GPT 5.4 as an LLM judge to score the binary correctnesses of candidate answers for a given question and reference answer.
For cases in which the LLM judge produced a refusal to judge, we performed judgments by hand.

To verify the quality of the LLM judgments as well as to check for idiosyncracies in the contrast sets, one author of this paper conducted a manual audit of a sample of full-set and contrast-set data from TriviaQA.
We sampled 50 questions from a contrast set and 50 questions from a non-contrast set (25 correct and 25 incorrect).
For each question, we checked whether the reference answer was correct by finding an online source.
If the reference answer was correct, we annotated whether it was ``incomplete'', i.e.\ other answers should also be acceptable even though they were not or would not be judged correct.
The auditor agreed with the LLM judgment in every case excluding incomplete cases (which are due to ambiguities with the underlying dataset), showing high reliability of LLM judgments.
The contrast set and non-contrast set had 50/50 and 47/50 correct reference answers, respectively.
All 3 questions in the non-contrast set that had incorrect reference answers were answered incorrectly.
The contrast set and non-contrast had 3/50 and 2/50 incomplete reference answers, respectively.
This confirms that the contrast set has similarly high quality data to the full set.
In fact, this audit suggests that the contrast set has even higher quality than the full set because the fact that at least one checkpoint answered a question correctly means that it is likely that the question was well-posed.

\subsection{Self-Consistency Implementation}
\label{app:sc-implementation}

We implement SC with the following design decisions to improve robustness.
Rather than sampling, we run beam search (beam size 10) and weight each beam by its sequence probability to efficiently cover the distribution of short-form generations \citep{fadeeva2026don}.
Beams are terminated at the first new line.
We use an off-the-shelf NLI model to group equivalent answers \citep{kuhn2023semantic}.\footnote{\url{https://huggingface.co/MoritzLaurer/DeBERTa-v3-base-mnli-fever-anli}}
Specifically, two answers are considered and equivalent and grouped together if each directed prediction has an argmax prediction of entailment (out of entailment, contradiction, and neutrality), following \citet{kuhn2023semantic}.
We define a candidate answer as a set of NLI-grouped beams, with generation probabilities summed.
We take the top (i.e.\ highest probability) candidate answer as the model's prediction, and its probability as the SC confidence.

For SC surrogate (\cref{sec:main-results}), since \C3 or \C4's predicted answer (i.e.\ top candidate answer) may not match \C2's predicted answer, we implement SC surrogate by using an NLI model to match \C3 or \C4's predicted answer against \C2's candidate answers, computing the confidence as the sum of the probabilities of the matching answers, or 0 if there is no match.

\subsection{LoRA Training Configuration}
\label{app:lora-config}

We train LoRA adapters \citep{hu2022lora} on all linear layers with $r=8$, $\alpha=16$, no dropout, no bias term, learning rate $2 \cdot 10^{-4}$, batch size of 16 questions, and one epoch.
We use the binary cross entropy loss function.
We schedule the learning rate to be stable at $2 \cdot 10^{-4}$ for the first 80\% of training and cosine-decay over the last 20\% to $2 \cdot 10^{-5}$.
We use the AdamW optimizer \citep{loshchilov2017decoupled} with $\beta_1 = 0.9$, $\beta_2 = 0.999$, $\epsilon = 10^{-8}$, weight decay 0.01.

We use the prompt below.
We compute the confidence as a probability-weighted average over the tokens \texttt{0}, \ldots, \texttt{9} (scaled to $[0, 1]$), similar to \citet{li2026conftuner}.
When applying a LoRA adapter to a model checkpoint, we use the checkpoint's token embedding matrix and token unembedding matrix (i.e.\ the LM head) without modification.
We found in early experiments that using the existing LM head rather than training and transferring a separate confidence head is crucial to achieve the forward transfer shown in \cref{fig:triviaqa-transfer,fig:jeopardy-transfer}.

\begin{tcolorbox}[colback=white!5!white, colframe=black!75!black, width=\columnwidth, title={Prompt for fine-tuning verbalized confidence.}]
\begin{PromptVerbatim}
Rate your confidence on a scale of 0-9, where 0 is completely uncertain and 9 is completely certain.

Question: \VarField{question}
Candidate answer: \VarField{candidate answer}
Confidence: 
\end{PromptVerbatim}
\end{tcolorbox}

\subsection{Significance Testing}
\label{app:sigtests}

We use a paired bootstrap test to check the statistical significance of the best reported result (bolded) outperforming another result (not bolded).
We first give background on how we compute the reported value of a metric, which we then build on to define our bootstrapping procedure.

\subsubsection{Background on the Reported Metric}

Suppose a metric $m$, confidence estimation method $f$, and base dataset $\tilde{\mathcal{D}}$.
The base dataset $\tilde{\mathcal{D}}$ consists of (question, reference answer) instances, but without candidate answers and correctness labels.
For a list of checkpoints $K$, each checkpoint $k \in K$ has an instantiation $\mathcal{D}_k$ of $\tilde{\mathcal{D}}$ with its own candidate answers and correctness labels.
Let $P$ be the set of unordered pairs of distinct checkpoints $p = (k, k') \in K \times K$ that have a sufficiently large contrast set (see \cref{app:agg-over-ckpts}).
For $p = (k, k') \in P$, we denote $\mathcal{D}_p$ as the concatenation of its two checkpoints' data $\mathcal{D}_{k}$ and $\mathcal{D}_{k'}$ (restricted to questions where their correctnesses differ, and with confidences normalized, as shown in \cref{fig:main}, right).
We define the concatenation $\mathcal{D} = \{\mathcal{D}_k\}_{k \in K}$ if $m$ is a full-set metric, and $\mathcal{D} = \{\mathcal{D}_p\}_{p \in P}$ if $m$ is a contrast-set metric.
We note that the data in $\mathcal{D}$ still carry information about the checkpoint identities $k$, as this is used in some metrics (\cref{app:agg-over-ckpts}).
To support stochastic confidence estimation methods, we suppose a set of random seeds $S$ and denote $f_s$ as the confidence estimator resulting from training or running $f$ with seed $s \in S$; deterministic confidence estimators simply ignore $s$.
The reported metric value is the metric value averaged over seeds:
\begin{align}
    m(f; \mathcal{D}) = \frac{1}{|S|} \sum_{s \in S} m(f_s; \mathcal{D})
    \label{eq:metric-avg-over-seeds}
\end{align}

\subsubsection{Paired Bootstrap Test}

We describe how we construct a bootstrap replicate $\mathcal{D}^b$ of $\mathcal{D}$.
We then use this procedure to construct $B = 10^4$ independent replicates $\{\mathcal{D}^b\}_{b \in [B]}$, giving us an empirical bootstrap distribution $\{m(f_{\rm best}; \mathcal{D}^b) - m(f; \mathcal{D}^b)\}_{b \in [B]}$ (with the sign flipped for metrics where lower is better).
We underline the result if the one-sided 95\% confidence interval contains 0.

We construct a given replicate in a way that accounts for data dependence and class stratification.
We have to account for data dependence because different checkpoints are evaluated using the same underlying questions $\tilde{\mathcal{D}}$, so treating the data in $\mathcal{D}$ as independent would inflate statistical significance.
We use class stratification so that each bootstrap replicate preserves certain statistics of the original data $\mathcal{D}$.
For full-set metrics, we use the classes [correct, incorrect], considered per checkpoint.
In other words, for each checkpoint, the number of correct instances and the number of incorrect instances should match between the replicate $\mathcal{D}^b$ and the original data $\mathcal{D}$.
As a corollary, the total number of instances for each checkpoint would match between $\mathcal{D}^b$ and $\mathcal{D}$.
For contrast-set metrics, we use the classes [improvement, regression] (i.e., the earlier checkpoint is incorrect and the later checkpoint is correct, or the earlier checkpoint is correct and the later checkpoint is incorrect), considered per checkpoint pair.
In other words, for each checkpoint pair, the number of improvement instances and the number of regression instances should match between the replicate $\mathcal{D}^b$ and the original data $\mathcal{D}$.
As a corollary, the total number of contrast-set instances for each checkpoint pair would match between $\mathcal{D}^b$ and $\mathcal{D}$.
Thus, a replicate in the full-set case has $2 \cdot |K|$ constraints ($K$ is the set of checkpoints), and a replicate in the contrast-set case has $2 \cdot |P|$ constraints ($P$ is the set of checkpoint pairs).

For ease of presentation, we assume the full-set case of $\mathcal{D} = \{\mathcal{D}_k\}_{k \in K}$ when describing our procedure for constructing a replicate; the contrast-set case is analogous by replacing $K$ with $P$.
We sample an infinite sequence of questions $Q^b = \{q^b_i\}_{i=1}^{\infty}$, sampled independently and with replacement from the full set of questions $\tilde{\mathcal{D}}$.
We construct a replicate $\mathcal{D}_k^b$ for each $\mathcal{D}_k$ and then concatenate the results to get $\mathcal{D}^b = \{\mathcal{D}^b_k\}_{k \in K}$.
We model dependence on a shared data pool by deriving $\mathcal{D}_k^b$ for different $k \in K$ from the same sequence $Q^b$.
For a given checkpoint $k$, each question $q_i^b$ belongs to class 1 (correct), class 2 (incorrect), or neither (only applicable to the contrast-set case).
To construct $\mathcal{D}_k^b$, we begin with an empty (multi)set and iterate through the sequence $Q^b$, adding a question if it belongs to class 1 or 2 and the class stratification quota for its class has not been met.
We stop once both quotas have been met.

To connect the interpretation of our constructed replicate $\mathcal{D}^b$ to the original data $\mathcal{D}$, we note that $\mathcal{D}$ can be viewed as being obtained from applying this procedure to a sequence of questions $Q$ that begins with the unique (duplicate-free) questions in $\tilde{\mathcal{D}}$.
Since this prefix fills all the quotas, the rest of $Q$ is ignored.

\subsection{Model Checkpoints}
\label{app:model-checkpoints}

For the checkpoints used in our experiments, \cref{tab:checkpoint-training-tokens} reports the checkpoint identities, training data sizes, and dataset accuracies.

\begin{table*}[t]
\caption{Cumulative training data and performance for the checkpoints used in our experiments.}
\label{tab:checkpoint-training-tokens}
\begin{center}
\small
\setlength{\tabcolsep}{5pt}
\begin{tabular}{clcccc}
\toprule
& & & \multicolumn{3}{c}{Accuracy (\%)} \\
\cmidrule(lr){4-6}
Short name & Checkpoint ID & \shortstack{Training data\\(trillions of tokens)} & TriviaQA & Jeopardy & BioASQ \\
\midrule
\multicolumn{6}{c}{\textit{Olmo 3 7B}} \\
\cmidrule(lr){1-6}
10\% & \texttt{stage1-step141000} & 0.59 & 51.9 & 64.8 & 40.2 \\
20\% & \texttt{stage1-step283000} & 1.19 & 55.5 & 68.5 & 41.7 \\
30\% & \texttt{stage1-step424000} & 1.78 & 57.8 & 70.7 & 44.0 \\
40\% & \texttt{stage1-step566000} & 2.37 & 59.5 & 72.1 & 44.5 \\
50\% & \texttt{stage1-step707000} & 2.97 & 60.8 & 73.6 & 47.7 \\
90\% & \texttt{stage1-step1272000} & 5.34 & 64.8 & 77.8 & 49.4 \\
100\% & \texttt{stage1-step1413814} & 5.93 & 65.6 & 78.6 & 51.1 \\
\midrule
\multicolumn{6}{c}{\textit{Marin 8B}} \\
\cmidrule(lr){1-6}
 & \texttt{kestrel} & 2.7 & 65.9 & 78.9 & 37.8 \\
 & \texttt{ocelot} & 3.78 & 67.1 & 79.5 & 38.9 \\
 & \texttt{jellyfish} & 4.78 & 67.0 & 81.0 & 50.5 \\
 & \texttt{phoenix} & 11.1 & 68.4 & 80.4 & 42.5 \\
 & \texttt{starling} & 12.4 & 76.1 & 85.9 & 54.1 \\
 & \texttt{deeper-starling} & 12.7 & 76.5 & 86.3 & 54.1 \\
\midrule
\multicolumn{6}{c}{\textit{Olmo 3 32B}} \\
\cmidrule(lr){1-6}
10\% & \texttt{stage1-step66000} & 0.55 & 62.7 & 75.4 & 46.9 \\
20\% & \texttt{stage1-step131000} & 1.10 & 67.3 & 80.0 & 50.3 \\
30\% & \texttt{stage1-step197000} & 1.65 & 68.5 & 81.8 & 52.2 \\
40\% & \texttt{stage1-step262000} & 2.20 & 69.7 & 83.7 & 53.8 \\
50\% & \texttt{stage1-step328000} & 2.75 & 71.2 & 84.2 & 53.5 \\
90\% & \texttt{stage1-step590120} & 4.95 & 76.8 & 87.6 & 59.5 \\
100\% & \texttt{stage1-step656000} & 5.50 & 77.7 & 88.8 & 61.5 \\
\bottomrule
\end{tabular}
\end{center}
\end{table*}

\subsection{Contrast Set Statistics}
\label{app:conset-stats}

Tables \ref{tab:conset-breakdown-triviaqa}, \ref{tab:conset-breakdown-jeopardy}, and \ref{tab:conset-breakdown-bioasq} report contrast set size statistics for TriviaQA, Jeopardy, and BioASQ.

\begin{table*}[ht]
\caption{TriviaQA contrast-set sizes for each pair of evaluation checkpoints.
The full-set size is 4960.
In a cell, the first number is the total size of the contrast set, and the second number is the number of questions that the row checkpoint gets incorrect and the column checkpoint gets correct.
Contrast-set sizes that are grayed out are too small (fewer than 500 questions) and are not used (\cref{app:agg-over-ckpts}).
}
\label{tab:conset-breakdown-triviaqa}
\begin{center}
\small
\setlength{\tabcolsep}{5pt}
\renewcommand{\arraystretch}{1.15}
\begin{tabular}{lcccc}
\toprule
\multicolumn{5}{c}{\textit{Olmo 3 7B}} \\
\cmidrule(lr){1-5}
& 40\% & 50\% & 90\% & 100\% \\
\midrule
40\% & -- & 642 (354) & 771 (516) & 753 (527) \\
50\% & 642 (288) & -- & 715 (455) & 697 (466) \\
90\% & 771 (255) & 715 (260) & -- & 538 (289) \\
100\% & 753 (226) & 697 (231) & 538 (249) & -- \\
\bottomrule
\end{tabular}
\vspace{5pt}
\begin{tabular}{lccc}
\multicolumn{4}{c}{\textit{Marin 8B}} \\
\cmidrule(lr){1-4}
& phoenix & starling & deeper\hbox{-}starling \\
\midrule
phoenix & -- & 698 (540) & 691 (547) \\
starling & 698 (158) & -- & \textcolor{gray}{147 (84)} \\
deeper\hbox{-}starling & 691 (144) & \textcolor{gray}{147 (63)} & -- \\
\bottomrule
\end{tabular}
\vspace{5pt}
\begin{tabular}{lcccc}
\multicolumn{5}{c}{\textit{Olmo 3 32B}} \\
\cmidrule(lr){1-5}
& 40\% & 50\% & 90\% & 100\% \\
\midrule
40\% & -- & 654 (364) & 676 (514) & 707 (551) \\
50\% & 654 (290) & -- & 660 (469) & 639 (480) \\
90\% & 676 (162) & 660 (191) & -- & \textcolor{gray}{467 (255)} \\
100\% & 707 (156) & 639 (159) & \textcolor{gray}{467 (212)} & -- \\
\bottomrule
\end{tabular}
\end{center}
\end{table*}

\begin{table*}[ht]
\caption{Jeopardy contrast-set sizes for each pair of evaluation checkpoints.
The full-set size is 4998.
In a cell, the first number is the total size of the contrast set, and the second number is the number of questions that the row checkpoint gets incorrect and the column checkpoint gets correct.
Contrast-set sizes that are grayed out are too small (fewer than 500 questions) and are not used (\cref{app:agg-over-ckpts}).}
\label{tab:conset-breakdown-jeopardy}
\begin{center}
\small
\setlength{\tabcolsep}{5pt}
\renewcommand{\arraystretch}{1.15}
\begin{tabular}{lcccc}
\toprule
\multicolumn{5}{c}{\textit{Olmo 3 7B}} \\
\cmidrule(lr){1-5}
& 40\% & 50\% & 90\% & 100\% \\
\midrule
40\% & -- & 738 (406) & 767 (526) & 796 (559) \\
50\% & 738 (332) & -- & 695 (453) & 720 (484) \\
90\% & 767 (241) & 695 (242) & -- & 517 (277) \\
100\% & 796 (237) & 720 (236) & 517 (240) & -- \\
\bottomrule
\end{tabular}
\vspace{5pt}
\begin{tabular}{lccc}
\multicolumn{4}{c}{\textit{Marin 8B}} \\
\cmidrule(lr){1-4}
& phoenix & starling & deeper\hbox{-}starling \\
\midrule
phoenix & -- & 604 (440) & 613 (454) \\
starling & 604 (164) & -- & \textcolor{gray}{153 (86)} \\
deeper\hbox{-}starling & 613 (159) & \textcolor{gray}{153 (67)} & -- \\
\bottomrule
\end{tabular}
\vspace{5pt}
\begin{tabular}{lcccc}
\multicolumn{5}{c}{\textit{Olmo 3 32B}} \\
\cmidrule(lr){1-5}
& 40\% & 50\% & 90\% & 100\% \\
\midrule
40\% & -- & 587 (307) & 613 (405) & 584 (421) \\
50\% & 587 (280) & -- & 546 (358) & 563 (397) \\
90\% & 613 (208) & 546 (188) & -- & \textcolor{gray}{405 (233)} \\
100\% & 584 (163) & 563 (166) & \textcolor{gray}{405 (172)} & -- \\
\bottomrule
\end{tabular}
\end{center}
\end{table*}

\begin{table*}[ht]
\caption{BioASQ contrast-set sizes for each pair of evaluation checkpoints.
The full-set size is 2719.
In a cell, the first number is the total size of the contrast set, and the second number is the number of questions that the row checkpoint gets incorrect and the column checkpoint gets correct.
Contrast-set sizes that are grayed out are too small (fewer than 300 questions) and are not used (\cref{app:agg-over-ckpts}).}
\label{tab:conset-breakdown-bioasq}
\begin{center}
\small
\setlength{\tabcolsep}{5pt}
\renewcommand{\arraystretch}{1.15}
\begin{tabular}{lcccc}
\toprule
\multicolumn{5}{c}{\textit{Olmo 3 7B}} \\
\cmidrule(lr){1-5}
& 40\% & 50\% & 90\% & 100\% \\
\midrule
40\% & -- & 457 (273) & 490 (312) & 475 (328) \\
50\% & 457 (184) & -- & 483 (264) & 466 (279) \\
90\% & 490 (178) & 483 (219) & -- & 389 (218) \\
100\% & 475 (147) & 466 (187) & 389 (171) & -- \\
\bottomrule
\end{tabular}
\vspace{5pt}
\begin{tabular}{lccc}
\multicolumn{4}{c}{\textit{Marin 8B}} \\
\cmidrule(lr){1-4}
& phoenix & starling & deeper\hbox{-}starling \\
\midrule
phoenix & -- & 564 (440) & 582 (449) \\
starling & 564 (124) & -- & \textcolor{gray}{172 (86)} \\
deeper\hbox{-}starling & 582 (133) & \textcolor{gray}{172 (86)} & -- \\
\bottomrule
\end{tabular}
\vspace{5pt}
\begin{tabular}{lcccc}
\multicolumn{5}{c}{\textit{Olmo 3 32B}} \\
\cmidrule(lr){1-5}
& 40\% & 50\% & 90\% & 100\% \\
\midrule
40\% & -- & 413 (202) & 486 (320) & 492 (351) \\
50\% & 413 (211) & -- & 451 (307) & 459 (339) \\
90\% & 486 (166) & 451 (144) & -- & 356 (206) \\
100\% & 492 (141) & 459 (120) & 356 (150) & -- \\
\bottomrule
\end{tabular}
\end{center}
\end{table*}

\section{Results and Analysis}
\label{app:results}

For tables in \cref{app:results} with bolded results, results are averaged over 3 seeds, and in each column, we bold the best result and underline results not significantly worse under a paired test ($\alpha = 0.05$; see \cref{app:sigtests} for tests).

\begin{table}[t]
\caption{Expanding on \cref{tab:oracle-vs-methods-abridged}, we consistently find that SC and non-oracle training fall short on contrast-set calibration compared to oracle training, even when full-set calibration is saturated. Notably, even after applying post-hoc calibration to SC (isotonic regression via the pool adjacent violators algorithm fitted on the train set of the non-oracle checkpoints, applied with linear interpolation), full-set ECE becomes negligible and yet contrast-set AUC does not improve.}
\label{tab:oracle-vs-methods-unabridged}
\begin{center}
\setlength{\tabcolsep}{8pt}
\small
\resizebox{\linewidth}{!}{%
\renewcommand{\arraystretch}{1.3}
\begin{tabular}{lcccccccccc}
\toprule
& \multicolumn{7}{c}{Contrast} & \multicolumn{3}{c}{Full} \\
\cmidrule(lr){2-8}\cmidrule(lr){9-11}
Method & $\Delta_0^b$ $\scriptstyle\uparrow$ & $\Delta_0$ $\scriptstyle\uparrow$ & $\Delta^b$ $\scriptstyle\uparrow$ & $\Delta$ $\scriptstyle\uparrow$ & AUC $\scriptstyle\uparrow$ & BS $\scriptstyle\downarrow$ & ECE $\scriptstyle\downarrow$ & AUC $\scriptstyle\uparrow$ & BS $\scriptstyle\downarrow$ & ECE $\scriptstyle\downarrow$ \\
\midrule
& \multicolumn{10}{c}{Olmo 3 7B, TriviaQA} \\
\cmidrule(lr){2-11}
End-correct baseline & 0.000 & 0.254 & 0.000 & 0.236 & 0.627 & 0.355 & 0.340 & 0.530 & 0.370 & 0.308 \\
Self-consistency & 0.297 & 0.329 & 0.190 & 0.212 & 0.733 & 0.212 & 0.084 & \underline{0.884} & 0.140 & 0.069 \\
\,\raisebox{0.5ex}{\(\llcorner\)} Post-hoc & 0.298 & 0.330 & 0.210 & 0.234 & 0.734 & 0.215 & 0.097 & \underline{0.884} & \underline{0.133} & \textbf{0.019} \\
Non-oracle training & 0.333 & 0.324 & 0.236 & 0.239 & 0.748 & 0.206 & 0.104 & 0.880 & 0.136 & 0.032 \\
Oracle training & \textbf{0.379} & \textbf{0.424} & \textbf{0.255} & \textbf{0.285} & \textbf{0.795} & \textbf{0.185} & \textbf{0.061} & \textbf{0.890} & \textbf{0.131} & 0.032 \\
\midrule
& \multicolumn{10}{c}{Olmo 3 7B, Jeopardy} \\
\cmidrule(lr){2-11}
End-correct baseline & 0.000 & 0.266 & 0.000 & 0.246 & 0.633 & 0.350 & 0.334 & 0.540 & 0.369 & 0.362 \\
Self-consistency & 0.333 & 0.360 & 0.201 & 0.221 & 0.741 & 0.209 & 0.087 & 0.873 & 0.120 & 0.020 \\
\,\raisebox{0.5ex}{\(\llcorner\)} Post-hoc & 0.333 & 0.360 & 0.215 & 0.236 & 0.742 & 0.211 & 0.092 & 0.873 & 0.119 & \textbf{0.016} \\
Non-oracle training & 0.476 & 0.476 & 0.360 & 0.365 & 0.821 & 0.177 & 0.094 & 0.871 & 0.119 & 0.038 \\
Oracle training & \textbf{0.506} & \textbf{0.536} & \textbf{0.379} & \textbf{0.402} & \textbf{0.855} & \textbf{0.156} & \textbf{0.057} & \textbf{0.888} & \textbf{0.111} & 0.020 \\
\midrule
& \multicolumn{10}{c}{Marin 8B, TriviaQA} \\
\cmidrule(lr){2-11}
End-correct baseline & 0.000 & \textbf{0.565} & 0.000 & \textbf{0.565} & 0.783 & 0.217 & 0.217 & 0.547 & 0.390 & 0.393 \\
Self-consistency & 0.263 & 0.482 & 0.196 & 0.349 & \underline{0.823} & \underline{0.172} & \underline{0.098} & \underline{0.897} & 0.115 & 0.052 \\
\,\raisebox{0.5ex}{\(\llcorner\)} Post-hoc & 0.263 & 0.482 & \underline{0.208} & 0.378 & \underline{0.825} & \underline{0.174} & \textbf{0.097} & \underline{0.897} & \underline{0.112} & \textbf{0.024} \\
Non-oracle training & 0.227 & 0.460 & 0.119 & 0.199 & 0.806 & 0.186 & 0.178 & 0.834 & 0.151 & 0.083 \\
Oracle training & \textbf{0.349} & 0.505 & \textbf{0.252} & 0.373 & \textbf{0.840} & \textbf{0.163} & \underline{0.105} & \textbf{0.901} & \textbf{0.110} & \underline{0.028} \\
\midrule
& \multicolumn{10}{c}{Marin 8B, Jeopardy} \\
\cmidrule(lr){2-11}
End-correct baseline & 0.000 & \underline{0.469} & 0.000 & \textbf{0.469} & 0.735 & 0.265 & 0.265 & 0.549 & 0.397 & 0.433 \\
Self-consistency & 0.331 & 0.377 & 0.215 & 0.255 & 0.760 & 0.202 & \underline{0.139} & \underline{0.886} & \underline{0.090} & 0.023 \\
\,\raisebox{0.5ex}{\(\llcorner\)} Post-hoc & 0.331 & 0.377 & 0.226 & 0.267 & 0.755 & 0.207 & \underline{0.136} & \underline{0.886} & \underline{0.089} & \textbf{0.015} \\
Non-oracle training & 0.397 & 0.394 & 0.137 & 0.139 & 0.759 & 0.205 & 0.200 & 0.834 & 0.124 & 0.129 \\
Oracle training & \textbf{0.459} & \textbf{0.503} & \textbf{0.350} & 0.377 & \textbf{0.825} & \textbf{0.175} & \textbf{0.123} & \textbf{0.890} & \textbf{0.088} & 0.027 \\
\midrule
& \multicolumn{10}{c}{Olmo 3 32B, TriviaQA} \\
\cmidrule(lr){2-11}
End-correct baseline & 0.000 & 0.423 & 0.000 & \textbf{0.390} & 0.712 & 0.273 & 0.248 & 0.548 & 0.364 & 0.350 \\
Self-consistency & \underline{0.362} & \underline{0.444} & \underline{0.264} & 0.327 & \underline{0.808} & 0.179 & \underline{0.098} & \underline{\textbf{0.903}} & 0.112 & 0.048 \\
\,\raisebox{0.5ex}{\(\llcorner\)} Post-hoc & \underline{0.363} & \underline{0.445} & \underline{0.274} & 0.342 & 0.806 & 0.183 & \underline{0.102} & \textbf{0.903} & \textbf{0.108} & \textbf{0.018} \\
Non-oracle training & 0.285 & 0.235 & 0.163 & 0.144 & 0.688 & 0.224 & 0.199 & 0.860 & 0.131 & 0.064 \\
Oracle training & \textbf{0.384} & \textbf{0.484} & \textbf{0.274} & 0.341 & \textbf{0.827} & \textbf{0.169} & \textbf{0.093} & \underline{0.898} & \underline{0.111} & 0.027 \\
\midrule
& \multicolumn{10}{c}{Olmo 3 32B, Jeopardy} \\
\cmidrule(lr){2-11}
End-correct baseline & 0.000 & 0.306 & 0.000 & 0.281 & 0.653 & 0.327 & 0.307 & 0.549 & 0.373 & 0.417 \\
Self-consistency & 0.320 & 0.364 & 0.234 & 0.259 & 0.754 & 0.207 & 0.103 & \underline{0.893} & \underline{0.081} & 0.020 \\
\,\raisebox{0.5ex}{\(\llcorner\)} Post-hoc & 0.320 & 0.364 & 0.246 & 0.274 & 0.755 & 0.209 & 0.106 & \underline{0.893} & \underline{0.081} & \textbf{0.018} \\
Non-oracle training & 0.399 & 0.374 & 0.228 & 0.218 & 0.757 & 0.201 & 0.140 & 0.860 & 0.092 & 0.063 \\
Oracle training & \textbf{0.475} & \textbf{0.513} & \textbf{0.342} & \textbf{0.370} & \textbf{0.840} & \textbf{0.163} & \textbf{0.075} & \textbf{0.894} & \textbf{0.081} & 0.029 \\
\bottomrule
\end{tabular}
}
\end{center}
\end{table}

\begin{table}[t]
\caption{Analogous setup to \cref{tab:oracle-vs-methods-unabridged}, but using a fitted normalization function (\cref{app:learn-condition-conset}) instead of \cref{eq:condition-conset-independent}. The two tables share the finding that SC and non-oracle training underperform oracle training on contrast-set calibration, even when full-set calibration is saturated.}
\label{tab:oracle-vs-methods-unabridged-learn1d}
\begin{center}
\setlength{\tabcolsep}{8pt}
\small
\resizebox{\linewidth}{!}{%
\renewcommand{\arraystretch}{1.3}
\begin{tabular}{lcccccccccc}
\toprule
& \multicolumn{7}{c}{Contrast} & \multicolumn{3}{c}{Full} \\
\cmidrule(lr){2-8}\cmidrule(lr){9-11}
Method & $\Delta_0^b$ $\scriptstyle\uparrow$ & $\Delta_0$ $\scriptstyle\uparrow$ & $\Delta^b$ $\scriptstyle\uparrow$ & $\Delta$ $\scriptstyle\uparrow$ & AUC $\scriptstyle\uparrow$ & BS $\scriptstyle\downarrow$ & ECE $\scriptstyle\downarrow$ & AUC $\scriptstyle\uparrow$ & BS $\scriptstyle\downarrow$ & ECE $\scriptstyle\downarrow$ \\
\midrule
& \multicolumn{10}{c}{Olmo 3 7B, TriviaQA} \\
\cmidrule(lr){2-11}
End-correct baseline & 0.000 & 0.254 & 0.000 & 0.080 & 0.627 & 0.230 & \textbf{0.000} & 0.530 & 0.370 & 0.308 \\
Self-consistency & 0.297 & 0.329 & 0.161 & 0.180 & 0.739 & 0.205 & 0.080 & \underline{0.884} & 0.140 & 0.069 \\
\,\raisebox{0.5ex}{\(\llcorner\)} Post-hoc & 0.298 & 0.330 & 0.163 & 0.181 & 0.740 & 0.205 & 0.080 & \underline{0.884} & \underline{0.133} & \textbf{0.019} \\
Non-oracle training & 0.333 & 0.324 & 0.211 & 0.219 & 0.759 & 0.196 & 0.092 & 0.880 & 0.136 & 0.032 \\
Oracle training & \textbf{0.379} & \textbf{0.424} & \textbf{0.254} & \textbf{0.283} & \textbf{0.803} & \textbf{0.180} & 0.056 & \textbf{0.890} & \textbf{0.131} & 0.032 \\
\midrule
& \multicolumn{10}{c}{Olmo 3 7B, Jeopardy} \\
\cmidrule(lr){2-11}
End-correct baseline & 0.000 & 0.266 & 0.000 & 0.088 & 0.633 & 0.228 & \textbf{0.000} & 0.540 & 0.369 & 0.362 \\
Self-consistency & 0.333 & 0.360 & 0.176 & 0.193 & 0.750 & 0.202 & 0.086 & 0.873 & 0.120 & 0.020 \\
\,\raisebox{0.5ex}{\(\llcorner\)} Post-hoc & 0.333 & 0.360 & 0.178 & 0.194 & 0.752 & 0.201 & 0.086 & 0.873 & 0.119 & \textbf{0.016} \\
Non-oracle training & 0.476 & 0.476 & 0.318 & 0.326 & 0.828 & 0.169 & 0.089 & 0.871 & 0.119 & 0.038 \\
Oracle training & \textbf{0.506} & \textbf{0.536} & \textbf{0.371} & \textbf{0.393} & \textbf{0.860} & \textbf{0.152} & 0.059 & \textbf{0.888} & \textbf{0.111} & 0.020 \\
\midrule
& \multicolumn{10}{c}{Marin 8B, TriviaQA} \\
\cmidrule(lr){2-11}
End-correct baseline & 0.000 & \textbf{0.565} & 0.000 & 0.320 & 0.783 & 0.170 & \textbf{0.000} & 0.547 & 0.390 & 0.393 \\
Self-consistency & 0.263 & 0.482 & 0.185 & 0.321 & \underline{0.828} & \underline{0.169} & 0.110 & \underline{0.897} & 0.115 & 0.052 \\
\,\raisebox{0.5ex}{\(\llcorner\)} Post-hoc & 0.263 & 0.482 & 0.179 & 0.320 & \underline{0.827} & \underline{0.170} & 0.108 & \underline{0.897} & \underline{0.112} & \textbf{0.024} \\
Non-oracle training & 0.227 & 0.460 & 0.162 & \underline{0.340} & \underline{0.833} & \underline{0.166} & 0.080 & 0.834 & 0.151 & 0.083 \\
Oracle training & \textbf{0.349} & 0.505 & \textbf{0.248} & \textbf{0.361} & \textbf{0.847} & \textbf{0.159} & 0.112 & \textbf{0.901} & \textbf{0.110} & \underline{0.028} \\
\midrule
& \multicolumn{10}{c}{Marin 8B, Jeopardy} \\
\cmidrule(lr){2-11}
End-correct baseline & 0.000 & \underline{0.469} & 0.000 & 0.220 & 0.735 & 0.195 & \textbf{0.000} & 0.549 & 0.397 & 0.433 \\
Self-consistency & 0.331 & 0.377 & 0.191 & 0.226 & 0.771 & 0.193 & 0.152 & \underline{0.886} & \underline{0.090} & 0.023 \\
\,\raisebox{0.5ex}{\(\llcorner\)} Post-hoc & 0.331 & 0.377 & 0.192 & 0.223 & 0.770 & 0.194 & 0.157 & \underline{0.886} & \underline{0.089} & \textbf{0.015} \\
Non-oracle training & 0.397 & 0.394 & 0.203 & 0.256 & 0.790 & 0.186 & 0.130 & 0.834 & 0.124 & 0.129 \\
Oracle training & \textbf{0.459} & \textbf{0.503} & \textbf{0.306} & \textbf{0.331} & \textbf{0.831} & \textbf{0.167} & 0.136 & \textbf{0.890} & \textbf{0.088} & 0.027 \\
\midrule
& \multicolumn{10}{c}{Olmo 3 32B, TriviaQA} \\
\cmidrule(lr){2-11}
End-correct baseline & 0.000 & 0.423 & 0.000 & 0.205 & 0.712 & 0.199 & \textbf{0.000} & 0.548 & 0.364 & 0.350 \\
Self-consistency & \underline{0.362} & \underline{0.444} & 0.246 & 0.303 & \underline{0.814} & 0.174 & 0.100 & \underline{\textbf{0.903}} & 0.112 & 0.048 \\
\,\raisebox{0.5ex}{\(\llcorner\)} Post-hoc & \underline{0.363} & \underline{0.445} & 0.246 & 0.303 & \underline{0.814} & 0.174 & 0.101 & \textbf{0.903} & \textbf{0.108} & \textbf{0.018} \\
Non-oracle training & 0.285 & 0.235 & 0.144 & 0.143 & 0.700 & 0.215 & 0.184 & 0.860 & 0.131 & 0.064 \\
Oracle training & \textbf{0.384} & \textbf{0.484} & \textbf{0.277} & \textbf{0.345} & \textbf{0.832} & \textbf{0.164} & 0.082 & \underline{0.898} & \underline{0.111} & 0.027 \\
\midrule
& \multicolumn{10}{c}{Olmo 3 32B, Jeopardy} \\
\cmidrule(lr){2-11}
End-correct baseline & 0.000 & 0.306 & 0.000 & 0.113 & 0.653 & 0.222 & \textbf{0.000} & 0.549 & 0.373 & 0.417 \\
Self-consistency & 0.320 & 0.364 & 0.206 & 0.229 & 0.771 & 0.193 & 0.094 & \underline{0.893} & \underline{0.081} & 0.020 \\
\,\raisebox{0.5ex}{\(\llcorner\)} Post-hoc & 0.320 & 0.364 & 0.202 & 0.225 & 0.769 & 0.194 & 0.094 & \underline{0.893} & \underline{0.081} & \textbf{0.018} \\
Non-oracle training & 0.399 & 0.374 & 0.225 & 0.224 & 0.768 & 0.194 & 0.128 & 0.860 & 0.092 & 0.063 \\
Oracle training & \textbf{0.475} & \textbf{0.513} & \textbf{0.342} & \textbf{0.369} & \textbf{0.848} & \textbf{0.158} & 0.070 & \textbf{0.894} & \textbf{0.081} & 0.029 \\
\bottomrule
\end{tabular}
}
\end{center}
\end{table}

\begin{table}[t]
\caption{Expanded version of \cref{tab:methods-vs-ablations-triviaqa}, for both TriviaQA and Jeopardy. A confidence estimator must outperform its ablations for a nontrivial demonstration of persistent calibration. Based on contrast-set AUC, SC outperforms its ablations in 5/6 cases, while non-oracle training outperforms its ablations in only 3/6 cases.}
\label{tab:methods-vs-ablations-unabridged}
\begin{center}
\setlength{\tabcolsep}{8pt}
\small
\resizebox{\linewidth}{!}{%
\renewcommand{\arraystretch}{1.3}
\begin{tabular}{lcccccccccc}
\toprule
& \multicolumn{7}{c}{Contrast} & \multicolumn{3}{c}{Full} \\
\cmidrule(lr){2-8}\cmidrule(lr){9-11}
Method & $\Delta_0^b$ $\scriptstyle\uparrow$ & $\Delta_0$ $\scriptstyle\uparrow$ & $\Delta^b$ $\scriptstyle\uparrow$ & $\Delta$ $\scriptstyle\uparrow$ & AUC $\scriptstyle\uparrow$ & BS $\scriptstyle\downarrow$ & ECE $\scriptstyle\downarrow$ & AUC $\scriptstyle\uparrow$ & BS $\scriptstyle\downarrow$ & ECE $\scriptstyle\downarrow$ \\
\midrule
& \multicolumn{10}{c}{Olmo 3 7B, TriviaQA} \\
\cmidrule(lr){2-11}
Self-consistency & \textbf{0.297} & \textbf{0.329} & \underline{0.190} & \textbf{0.212} & \textbf{0.733} & \textbf{0.212} & \textbf{0.084} & \textbf{0.884} & \textbf{0.140} & 0.069 \\
\,\raisebox{0.5ex}{\(\llcorner\)} Surrogate & 0.211 & 0.191 & \textbf{0.204} & \underline{0.184} & 0.637 & 0.304 & 0.208 & 0.859 & 0.163 & 0.094 \\
\,\raisebox{0.5ex}{\(\llcorner\)} Copy & 0.000 & 0.000 & 0.000 & 0.000 & 0.500 & 0.250 & 0.127 & 0.832 & 0.161 & \textbf{0.033} \\
\midrule
\addlinespace[3pt]
Non-oracle training & \textbf{0.333} & \textbf{0.324} & \textbf{0.236} & \textbf{0.239} & \textbf{0.748} & \textbf{0.206} & \underline{0.104} & \textbf{0.880} & \textbf{0.136} & \textbf{0.032} \\
\,\raisebox{0.5ex}{\(\llcorner\)} Surrogate & \underline{0.318} & \underline{0.323} & 0.182 & 0.186 & 0.729 & \underline{0.210} & \textbf{0.103} & 0.869 & 0.146 & 0.060 \\
\,\raisebox{0.5ex}{\(\llcorner\)} Copy & 0.000 & 0.000 & 0.000 & 0.000 & 0.500 & 0.250 & 0.127 & 0.839 & 0.163 & 0.064 \\
\midrule
& \multicolumn{10}{c}{Olmo 3 7B, Jeopardy} \\
\cmidrule(lr){2-11}
Self-consistency & \textbf{0.333} & \textbf{0.360} & 0.201 & 0.221 & \textbf{0.741} & \textbf{0.209} & \textbf{0.087} & \textbf{0.873} & \textbf{0.120} & \textbf{0.020} \\
\,\raisebox{0.5ex}{\(\llcorner\)} Surrogate & \underline{0.309} & 0.302 & \textbf{0.280} & \textbf{0.271} & 0.700 & 0.261 & 0.176 & \underline{0.869} & 0.154 & 0.138 \\
\,\raisebox{0.5ex}{\(\llcorner\)} Copy & 0.000 & 0.000 & 0.000 & 0.000 & 0.500 & 0.250 & 0.133 & 0.815 & 0.146 & 0.046 \\
\midrule
\addlinespace[3pt]
Non-oracle training & \textbf{0.476} & \textbf{0.476} & \textbf{0.360} & \textbf{0.365} & \textbf{0.821} & \textbf{0.177} & \textbf{0.094} & \textbf{0.871} & \textbf{0.119} & \textbf{0.038} \\
\,\raisebox{0.5ex}{\(\llcorner\)} Surrogate & \underline{0.444} & 0.437 & 0.317 & 0.312 & 0.805 & \underline{0.181} & 0.102 & 0.864 & 0.127 & 0.051 \\
\,\raisebox{0.5ex}{\(\llcorner\)} Copy & 0.000 & 0.000 & 0.000 & 0.000 & 0.500 & 0.250 & 0.133 & 0.800 & 0.158 & 0.071 \\
\midrule
& \multicolumn{10}{c}{Marin 8B, TriviaQA} \\
\cmidrule(lr){2-11}
Self-consistency & \underline{0.263} & \textbf{0.482} & 0.196 & \textbf{0.349} & \textbf{0.823} & \textbf{0.172} & \textbf{0.098} & \textbf{0.897} & \textbf{0.115} & \textbf{0.052} \\
\,\raisebox{0.5ex}{\(\llcorner\)} Surrogate & \textbf{0.277} & 0.283 & \textbf{0.268} & 0.266 & 0.695 & 0.275 & 0.224 & 0.871 & 0.161 & 0.127 \\
\,\raisebox{0.5ex}{\(\llcorner\)} Copy & 0.000 & 0.000 & 0.000 & 0.000 & 0.500 & 0.250 & 0.283 & 0.829 & 0.146 & \underline{0.053} \\
\midrule
\addlinespace[3pt]
Non-oracle training & 0.227 & \textbf{0.460} & 0.119 & 0.199 & \textbf{0.806} & \textbf{0.186} & \textbf{0.178} & 0.834 & 0.151 & \underline{0.083} \\
\,\raisebox{0.5ex}{\(\llcorner\)} Surrogate & \textbf{0.318} & 0.339 & \textbf{0.225} & \textbf{0.239} & 0.748 & 0.206 & 0.207 & \textbf{0.866} & \textbf{0.140} & \textbf{0.082} \\
\,\raisebox{0.5ex}{\(\llcorner\)} Copy & 0.000 & 0.000 & 0.000 & 0.000 & 0.500 & 0.250 & 0.283 & 0.830 & 0.159 & 0.092 \\
\midrule
& \multicolumn{10}{c}{Marin 8B, Jeopardy} \\
\cmidrule(lr){2-11}
Self-consistency & 0.331 & 0.377 & 0.215 & 0.255 & 0.760 & \textbf{0.202} & \textbf{0.139} & 0.886 & \textbf{0.090} & \textbf{0.023} \\
\,\raisebox{0.5ex}{\(\llcorner\)} Surrogate & \textbf{0.515} & \textbf{0.478} & \textbf{0.448} & \textbf{0.425} & \textbf{0.806} & \underline{0.203} & 0.182 & \textbf{0.896} & 0.122 & 0.126 \\
\,\raisebox{0.5ex}{\(\llcorner\)} Copy & 0.000 & 0.000 & 0.000 & 0.000 & 0.500 & 0.250 & 0.235 & 0.831 & 0.112 & 0.053 \\
\midrule
\addlinespace[3pt]
Non-oracle training & 0.397 & 0.394 & 0.137 & 0.139 & 0.759 & 0.205 & 0.200 & 0.834 & 0.124 & 0.129 \\
\,\raisebox{0.5ex}{\(\llcorner\)} Surrogate & \textbf{0.504} & \textbf{0.496} & \textbf{0.400} & \textbf{0.388} & \textbf{0.843} & \textbf{0.163} & \textbf{0.154} & \textbf{0.877} & \textbf{0.098} & \textbf{0.048} \\
\,\raisebox{0.5ex}{\(\llcorner\)} Copy & 0.000 & 0.000 & 0.000 & 0.000 & 0.500 & 0.250 & 0.235 & 0.806 & 0.128 & 0.074 \\
\midrule
& \multicolumn{10}{c}{Olmo 3 32B, TriviaQA} \\
\cmidrule(lr){2-11}
Self-consistency & \textbf{0.362} & \textbf{0.444} & \underline{0.264} & \textbf{0.327} & \textbf{0.808} & \textbf{0.179} & \textbf{0.098} & \textbf{0.903} & \textbf{0.112} & 0.048 \\
\,\raisebox{0.5ex}{\(\llcorner\)} Surrogate & 0.288 & 0.263 & \textbf{0.279} & 0.253 & 0.688 & 0.274 & 0.212 & 0.883 & 0.144 & 0.107 \\
\,\raisebox{0.5ex}{\(\llcorner\)} Copy & 0.000 & 0.000 & 0.000 & 0.000 & 0.500 & 0.250 & 0.212 & 0.846 & 0.137 & \textbf{0.040} \\
\midrule
\addlinespace[3pt]
Non-oracle training & \underline{0.285} & 0.235 & \textbf{0.163} & 0.144 & 0.688 & 0.224 & 0.199 & \textbf{0.860} & \textbf{0.131} & \textbf{0.064} \\
\,\raisebox{0.5ex}{\(\llcorner\)} Surrogate & \textbf{0.290} & \textbf{0.296} & \underline{0.157} & \textbf{0.161} & \textbf{0.716} & \textbf{0.215} & \textbf{0.175} & \underline{0.858} & 0.136 & \underline{0.068} \\
\,\raisebox{0.5ex}{\(\llcorner\)} Copy & 0.000 & 0.000 & 0.000 & 0.000 & 0.500 & 0.250 & 0.212 & 0.832 & 0.148 & 0.070 \\
\midrule
& \multicolumn{10}{c}{Olmo 3 32B, Jeopardy} \\
\cmidrule(lr){2-11}
Self-consistency & \textbf{0.320} & \textbf{0.364} & 0.234 & \underline{0.259} & \textbf{0.754} & \textbf{0.207} & \textbf{0.103} & \textbf{0.893} & \textbf{0.081} & \textbf{0.020} \\
\,\raisebox{0.5ex}{\(\llcorner\)} Surrogate & \underline{0.312} & 0.298 & \textbf{0.284} & \textbf{0.272} & 0.704 & 0.271 & 0.200 & \underline{0.889} & 0.115 & 0.117 \\
\,\raisebox{0.5ex}{\(\llcorner\)} Copy & 0.000 & 0.000 & 0.000 & 0.000 & 0.500 & 0.250 & 0.153 & 0.834 & 0.102 & 0.051 \\
\midrule
\addlinespace[3pt]
Non-oracle training & \underline{0.399} & \underline{0.374} & 0.228 & 0.218 & \underline{0.757} & \underline{0.201} & \underline{0.140} & \underline{0.860} & \textbf{0.092} & 0.063 \\
\,\raisebox{0.5ex}{\(\llcorner\)} Surrogate & \textbf{0.417} & \textbf{0.398} & \textbf{0.262} & \textbf{0.251} & \textbf{0.764} & \textbf{0.200} & \textbf{0.133} & \textbf{0.866} & 0.095 & \textbf{0.056} \\
\,\raisebox{0.5ex}{\(\llcorner\)} Copy & 0.000 & 0.000 & 0.000 & 0.000 & 0.500 & 0.250 & 0.153 & 0.817 & 0.113 & 0.064 \\
\bottomrule
\end{tabular}
}
\end{center}
\end{table}

\begin{table}[t]
\caption{Multi-checkpoint and multi-answer training often yields substantial gains. TriviaQA results; see \cref{tab:mvs-ckpt-ans-jeopardy} for Jeopardy.}
\label{tab:mvs-ckpt-ans-triviaqa}
\begin{center}
\setlength{\tabcolsep}{8pt}
\small
\resizebox{\linewidth}{!}{%
\renewcommand{\arraystretch}{1.3}
\begin{tabular}{lcccccccccc}
\toprule
& \multicolumn{7}{c}{Contrast} & \multicolumn{3}{c}{Full} \\
\cmidrule(lr){2-8}\cmidrule(lr){9-11}
Method & $\Delta_0^b$ $\scriptstyle\uparrow$ & $\Delta_0$ $\scriptstyle\uparrow$ & $\Delta^b$ $\scriptstyle\uparrow$ & $\Delta$ $\scriptstyle\uparrow$ & AUC $\scriptstyle\uparrow$ & BS $\scriptstyle\downarrow$ & ECE $\scriptstyle\downarrow$ & AUC $\scriptstyle\uparrow$ & BS $\scriptstyle\downarrow$ & ECE $\scriptstyle\downarrow$ \\
\midrule
\multicolumn{1}{l}{\:\textit{Non-oracle}} & \multicolumn{10}{c}{Olmo 3 7B} \\
\cmidrule(lr){1-1}
\cmidrule(lr){2-11}
Multi-ckpt, $\text{n}_{\text{cand}}$=2 & \textbf{0.333} & 0.324 & \textbf{0.236} & \textbf{0.239} & \underline{0.748} & 0.206 & 0.104 & \textbf{0.880} & \textbf{0.136} & \underline{0.032} \\
Multi-ckpt, $\text{n}_{\text{cand}}$=1 & \underline{0.332} & \underline{0.335} & 0.223 & 0.233 & \underline{0.749} & \underline{0.204} & 0.094 & 0.877 & 0.137 & \textbf{0.032} \\
Single-ckpt, $\text{n}_{\text{cand}}$=2 & \underline{0.332} & \textbf{0.340} & 0.216 & 0.227 & \textbf{0.750} & \textbf{0.203} & 0.093 & 0.876 & 0.140 & 0.048 \\
Single-ckpt, $\text{n}_{\text{cand}}$=1 & 0.314 & \underline{0.339} & 0.208 & 0.225 & \underline{0.746} & 0.206 & \textbf{0.086} & 0.870 & 0.146 & 0.063 \\
\midrule
\addlinespace[3pt]
\multicolumn{1}{l}{\:\textit{Oracle}} & \multicolumn{10}{c}{} \\
\cmidrule(lr){1-1}
Multi-ckpt, $\text{n}_{\text{cand}}$=2 & \textbf{0.379} & \textbf{0.424} & \textbf{0.255} & \textbf{0.285} & \textbf{0.795} & \textbf{0.185} & 0.061 & \textbf{0.890} & \textbf{0.131} & 0.032 \\
Multi-ckpt, $\text{n}_{\text{cand}}$=1 & 0.356 & \underline{0.414} & 0.239 & 0.274 & 0.790 & 0.187 & \textbf{0.058} & 0.887 & 0.132 & \textbf{0.026} \\
Resp-ckpt, $\text{n}_{\text{cand}}$=2 & 0.351 & 0.396 & 0.241 & 0.274 & 0.778 & 0.193 & 0.070 & 0.884 & 0.133 & 0.030 \\
Resp-ckpt, $\text{n}_{\text{cand}}$=1 & 0.317 & 0.357 & 0.219 & 0.246 & 0.756 & 0.203 & 0.083 & 0.880 & 0.136 & 0.033 \\
\midrule
\multicolumn{1}{l}{\:\textit{Non-oracle}} & \multicolumn{10}{c}{Marin 8B} \\
\cmidrule(lr){1-1}
\cmidrule(lr){2-11}
Multi-ckpt, $\text{n}_{\text{cand}}$=2 & \underline{0.227} & 0.460 & \textbf{0.119} & 0.199 & 0.806 & 0.186 & 0.178 & \textbf{0.834} & \textbf{0.151} & \textbf{0.083} \\
Multi-ckpt, $\text{n}_{\text{cand}}$=1 & \textbf{0.229} & \textbf{0.485} & 0.112 & \textbf{0.207} & \textbf{0.815} & \textbf{0.182} & \textbf{0.167} & 0.831 & 0.153 & \underline{0.084} \\
Single-ckpt, $\text{n}_{\text{cand}}$=2 & \underline{0.221} & 0.213 & 0.050 & 0.044 & 0.642 & 0.237 & 0.274 & 0.783 & 0.175 & 0.086 \\
Single-ckpt, $\text{n}_{\text{cand}}$=1 & \underline{0.213} & 0.212 & 0.047 & 0.045 & 0.647 & 0.236 & 0.272 & 0.780 & 0.175 & \underline{0.085} \\
\midrule
\addlinespace[3pt]
\multicolumn{1}{l}{\:\textit{Oracle}} & \multicolumn{10}{c}{} \\
\cmidrule(lr){1-1}
Multi-ckpt, $\text{n}_{\text{cand}}$=2 & \textbf{0.349} & \textbf{0.505} & \underline{0.252} & \underline{0.373} & \textbf{0.840} & \textbf{0.163} & \textbf{0.105} & \underline{0.901} & 0.110 & 0.028 \\
Multi-ckpt, $\text{n}_{\text{cand}}$=1 & 0.315 & \underline{0.490} & 0.232 & 0.353 & 0.835 & 0.166 & 0.110 & 0.897 & 0.111 & \underline{0.023} \\
Resp-ckpt, $\text{n}_{\text{cand}}$=2 & \underline{0.330} & \underline{0.496} & \textbf{0.257} & \textbf{0.375} & \underline{0.837} & \underline{0.166} & \underline{0.105} & \textbf{0.901} & \textbf{0.110} & 0.025 \\
Resp-ckpt, $\text{n}_{\text{cand}}$=1 & \underline{0.324} & 0.466 & 0.244 & 0.351 & 0.823 & 0.173 & 0.121 & 0.896 & 0.111 & \textbf{0.021} \\
\midrule
\multicolumn{1}{l}{\:\textit{Non-oracle}} & \multicolumn{10}{c}{Olmo 3 32B} \\
\cmidrule(lr){1-1}
\cmidrule(lr){2-11}
Multi-ckpt, $\text{n}_{\text{cand}}$=2 & \textbf{0.285} & \textbf{0.235} & \textbf{0.163} & \textbf{0.144} & \textbf{0.688} & \textbf{0.224} & 0.199 & \textbf{0.860} & \textbf{0.131} & 0.064 \\
Multi-ckpt, $\text{n}_{\text{cand}}$=1 & 0.268 & 0.221 & 0.152 & 0.137 & 0.683 & 0.225 & \textbf{0.198} & \underline{0.860} & 0.132 & \textbf{0.063} \\
Single-ckpt, $\text{n}_{\text{cand}}$=2 & 0.251 & 0.208 & 0.140 & 0.125 & 0.673 & 0.228 & 0.201 & 0.849 & 0.138 & 0.070 \\
Single-ckpt, $\text{n}_{\text{cand}}$=1 & 0.225 & 0.180 & 0.119 & 0.103 & 0.648 & 0.236 & 0.214 & 0.836 & 0.143 & 0.072 \\
\midrule
\addlinespace[3pt]
\multicolumn{1}{l}{\:\textit{Oracle}} & \multicolumn{10}{c}{} \\
\cmidrule(lr){1-1}
Multi-ckpt, $\text{n}_{\text{cand}}$=2 & \textbf{0.384} & \textbf{0.484} & \textbf{0.274} & \textbf{0.341} & \textbf{0.827} & \textbf{0.169} & 0.093 & \textbf{0.898} & \textbf{0.111} & 0.027 \\
Multi-ckpt, $\text{n}_{\text{cand}}$=1 & 0.345 & \underline{0.478} & 0.249 & 0.330 & 0.824 & \underline{0.170} & \textbf{0.085} & 0.892 & 0.114 & 0.028 \\
Resp-ckpt, $\text{n}_{\text{cand}}$=2 & 0.352 & 0.459 & 0.258 & 0.327 & 0.812 & 0.176 & 0.097 & 0.892 & 0.113 & \textbf{0.024} \\
Resp-ckpt, $\text{n}_{\text{cand}}$=1 & 0.324 & 0.459 & 0.226 & 0.315 & 0.808 & 0.178 & \underline{0.087} & 0.884 & 0.118 & 0.027 \\
\bottomrule
\end{tabular}
}
\end{center}
\end{table}

\begin{table}[t]
\caption{Multi-checkpoint and multi-answer training often yields substantial gains. Jeopardy results; see \cref{tab:mvs-ckpt-ans-triviaqa} for TriviaQA.}
\label{tab:mvs-ckpt-ans-jeopardy}
\begin{center}
\setlength{\tabcolsep}{8pt}
\small
\resizebox{\linewidth}{!}{%
\renewcommand{\arraystretch}{1.3}
\begin{tabular}{lcccccccccc}
\toprule
& \multicolumn{7}{c}{Contrast} & \multicolumn{3}{c}{Full} \\
\cmidrule(lr){2-8}\cmidrule(lr){9-11}
Method & $\Delta_0^b$ $\scriptstyle\uparrow$ & $\Delta_0$ $\scriptstyle\uparrow$ & $\Delta^b$ $\scriptstyle\uparrow$ & $\Delta$ $\scriptstyle\uparrow$ & AUC $\scriptstyle\uparrow$ & BS $\scriptstyle\downarrow$ & ECE $\scriptstyle\downarrow$ & AUC $\scriptstyle\uparrow$ & BS $\scriptstyle\downarrow$ & ECE $\scriptstyle\downarrow$ \\
\midrule
\multicolumn{1}{l}{\:\textit{Non-oracle}} & \multicolumn{10}{c}{Olmo 3 7B} \\
\cmidrule(lr){1-1}
\cmidrule(lr){2-11}
Multi-ckpt, $\text{n}_{\text{cand}}$=2 & \textbf{0.476} & \textbf{0.476} & \textbf{0.360} & \textbf{0.365} & \textbf{0.821} & \textbf{0.177} & \underline{0.094} & \textbf{0.871} & \textbf{0.119} & \textbf{0.038} \\
Multi-ckpt, $\text{n}_{\text{cand}}$=1 & 0.438 & 0.438 & 0.327 & 0.328 & 0.802 & 0.186 & 0.103 & 0.863 & 0.124 & 0.042 \\
Single-ckpt, $\text{n}_{\text{cand}}$=2 & 0.447 & 0.457 & 0.318 & 0.327 & 0.808 & 0.181 & \underline{0.094} & 0.860 & 0.124 & \underline{0.039} \\
Single-ckpt, $\text{n}_{\text{cand}}$=1 & 0.418 & 0.441 & 0.278 & 0.294 & 0.796 & 0.185 & \textbf{0.090} & 0.851 & 0.130 & 0.050 \\
\midrule
\addlinespace[3pt]
\multicolumn{1}{l}{\:\textit{Oracle}} & \multicolumn{10}{c}{} \\
\cmidrule(lr){1-1}
Multi-ckpt, $\text{n}_{\text{cand}}$=2 & \textbf{0.506} & \textbf{0.536} & \textbf{0.379} & \textbf{0.402} & \textbf{0.855} & \textbf{0.156} & \underline{0.057} & \textbf{0.888} & \textbf{0.111} & \textbf{0.020} \\
Multi-ckpt, $\text{n}_{\text{cand}}$=1 & 0.478 & 0.518 & 0.343 & 0.372 & 0.845 & 0.161 & \textbf{0.055} & 0.880 & 0.117 & 0.033 \\
Resp-ckpt, $\text{n}_{\text{cand}}$=2 & 0.471 & 0.485 & 0.352 & 0.364 & 0.826 & 0.172 & 0.079 & 0.877 & 0.116 & 0.028 \\
Resp-ckpt, $\text{n}_{\text{cand}}$=1 & 0.436 & 0.456 & 0.311 & 0.326 & 0.806 & 0.182 & 0.087 & 0.864 & 0.123 & 0.038 \\
\midrule
\multicolumn{1}{l}{\:\textit{Non-oracle}} & \multicolumn{10}{c}{Marin 8B} \\
\cmidrule(lr){1-1}
\cmidrule(lr){2-11}
Multi-ckpt, $\text{n}_{\text{cand}}$=2 & \textbf{0.397} & \underline{0.394} & \textbf{0.137} & 0.139 & 0.759 & 0.205 & 0.200 & \textbf{0.834} & 0.124 & 0.129 \\
Multi-ckpt, $\text{n}_{\text{cand}}$=1 & 0.318 & \textbf{0.398} & 0.120 & \textbf{0.146} & \textbf{0.776} & \textbf{0.200} & \textbf{0.178} & 0.826 & \textbf{0.121} & \textbf{0.111} \\
Single-ckpt, $\text{n}_{\text{cand}}$=2 & 0.240 & 0.076 & 0.059 & 0.021 & 0.562 & 0.248 & 0.261 & 0.757 & 0.152 & 0.176 \\
Single-ckpt, $\text{n}_{\text{cand}}$=1 & 0.235 & 0.092 & 0.054 & 0.023 & 0.571 & 0.246 & 0.256 & 0.756 & 0.149 & 0.167 \\
\midrule
\addlinespace[3pt]
\multicolumn{1}{l}{\:\textit{Oracle}} & \multicolumn{10}{c}{} \\
\cmidrule(lr){1-1}
Multi-ckpt, $\text{n}_{\text{cand}}$=2 & \textbf{0.459} & \textbf{0.503} & 0.350 & \underline{0.377} & \textbf{0.825} & \underline{0.175} & 0.123 & 0.890 & 0.088 & 0.027 \\
Multi-ckpt, $\text{n}_{\text{cand}}$=1 & 0.424 & \underline{0.496} & 0.310 & 0.365 & \underline{\textbf{0.825}} & \textbf{0.174} & \textbf{0.105} & 0.881 & 0.089 & \textbf{0.020} \\
Resp-ckpt, $\text{n}_{\text{cand}}$=2 & \underline{0.453} & \underline{0.492} & \textbf{0.360} & \textbf{0.382} & \underline{0.824} & \underline{0.176} & 0.126 & \textbf{0.891} & \textbf{0.087} & 0.027 \\
Resp-ckpt, $\text{n}_{\text{cand}}$=1 & 0.410 & 0.479 & 0.306 & 0.354 & \underline{0.818} & \underline{0.178} & 0.115 & 0.880 & 0.089 & 0.022 \\
\midrule
\multicolumn{1}{l}{\:\textit{Non-oracle}} & \multicolumn{10}{c}{Olmo 3 32B} \\
\cmidrule(lr){1-1}
\cmidrule(lr){2-11}
Multi-ckpt, $\text{n}_{\text{cand}}$=2 & \textbf{0.399} & \textbf{0.374} & \textbf{0.228} & \textbf{0.218} & \textbf{0.757} & \textbf{0.201} & 0.140 & \textbf{0.860} & \underline{0.092} & 0.063 \\
Multi-ckpt, $\text{n}_{\text{cand}}$=1 & 0.363 & \underline{0.369} & 0.200 & 0.204 & \underline{0.755} & \underline{0.202} & 0.131 & 0.853 & \textbf{0.092} & \textbf{0.053} \\
Single-ckpt, $\text{n}_{\text{cand}}$=2 & 0.315 & 0.318 & 0.174 & 0.178 & 0.728 & 0.212 & 0.140 & 0.830 & 0.100 & 0.073 \\
Single-ckpt, $\text{n}_{\text{cand}}$=1 & 0.310 & 0.339 & 0.164 & 0.181 & 0.734 & 0.210 & \textbf{0.125} & 0.825 & 0.099 & 0.061 \\
\midrule
\addlinespace[3pt]
\multicolumn{1}{l}{\:\textit{Oracle}} & \multicolumn{10}{c}{} \\
\cmidrule(lr){1-1}
Multi-ckpt, $\text{n}_{\text{cand}}$=2 & \textbf{0.475} & \textbf{0.513} & \textbf{0.342} & \textbf{0.370} & \textbf{0.840} & \textbf{0.163} & 0.075 & \textbf{0.894} & \textbf{0.081} & 0.029 \\
Multi-ckpt, $\text{n}_{\text{cand}}$=1 & 0.414 & 0.469 & 0.284 & 0.322 & 0.819 & 0.173 & \underline{0.067} & 0.879 & 0.084 & \textbf{0.021} \\
Resp-ckpt, $\text{n}_{\text{cand}}$=2 & 0.448 & 0.481 & 0.323 & 0.347 & 0.824 & 0.172 & 0.083 & 0.884 & 0.083 & 0.026 \\
Resp-ckpt, $\text{n}_{\text{cand}}$=1 & 0.375 & 0.450 & 0.254 & 0.305 & 0.805 & 0.181 & \textbf{0.067} & 0.866 & 0.087 & \underline{0.022} \\
\bottomrule
\end{tabular}
}
\end{center}
\end{table}

\begin{table}[t]
\caption{Counterpart to \cref{tab:multi-ckpt-vs-ablations-olmo7b-triviaqa} (TriviaQA) for Jeopardy. Multi-checkpoint training continues to outperform its two ablations, although the single-checkpoint data ablation is able to match the unablated multi-checkpoint training on the class-balanced contrast-set metrics $\Delta^b_0, \Delta^b$, suggesting that access to multiple checkpoints' weights accounts for most of the performance gain of multi-checkpoint training.}
\label{tab:multi-ckpt-vs-ablations-olmo7b-jeopardy}
\begin{center}
\setlength{\tabcolsep}{8pt}
\small
\resizebox{\linewidth}{!}{%
\renewcommand{\arraystretch}{1.3}
\begin{tabular}{lcccccccccc}
\toprule
& \multicolumn{7}{c}{Contrast} & \multicolumn{3}{c}{Full} \\
\cmidrule(lr){2-8}\cmidrule(lr){9-11}
Method & $\Delta_0^b$ $\scriptstyle\uparrow$ & $\Delta_0$ $\scriptstyle\uparrow$ & $\Delta^b$ $\scriptstyle\uparrow$ & $\Delta$ $\scriptstyle\uparrow$ & AUC $\scriptstyle\uparrow$ & BS $\scriptstyle\downarrow$ & ECE $\scriptstyle\downarrow$ & AUC $\scriptstyle\uparrow$ & BS $\scriptstyle\downarrow$ & ECE $\scriptstyle\downarrow$ \\
\midrule
Multi-ckpt & \underline{0.506} & \textbf{0.536} & \underline{0.379} & \textbf{0.402} & \textbf{0.855} & \textbf{0.156} & \textbf{0.057} & \textbf{0.888} & \textbf{0.111} & \textbf{0.020} \\
\,\raisebox{0.5ex}{\(\llcorner\)} Single-ckpt data & \textbf{0.517} & 0.517 & \textbf{0.382} & 0.383 & 0.841 & 0.164 & 0.087 & 0.885 & 0.115 & 0.041 \\
Resp-ckpt & 0.471 & 0.485 & 0.352 & 0.364 & 0.826 & 0.172 & 0.079 & 0.877 & 0.116 & 0.028 \\
\,\raisebox{0.5ex}{\(\llcorner\)} Pooled data & \underline{0.506} & 0.499 & \underline{0.380} & 0.376 & 0.833 & 0.169 & 0.094 & 0.884 & 0.113 & 0.027 \\
\bottomrule
\end{tabular}
}
\end{center}
\end{table}

\begin{table}[t]
\caption{GCMs underperform oracle training, especially for larger models, highlighting that a static external verifier does not adapt to a model's capability, motivating meta-knowledge for persistent calibration.
This holds even with post-hoc calibration on GCM; see \cref{tab:oracle-vs-methods-unabridged} for a description.}
\label{tab:oracle-vs-gcm-posthoc}
\begin{center}
\setlength{\tabcolsep}{8pt}
\small
\resizebox{\linewidth}{!}{%
\renewcommand{\arraystretch}{1.3}
\begin{tabular}{lcccccccccc}
\toprule
& \multicolumn{7}{c}{Contrast} & \multicolumn{3}{c}{Full} \\
\cmidrule(lr){2-8}\cmidrule(lr){9-11}
Method & $\Delta_0^b$ $\scriptstyle\uparrow$ & $\Delta_0$ $\scriptstyle\uparrow$ & $\Delta^b$ $\scriptstyle\uparrow$ & $\Delta$ $\scriptstyle\uparrow$ & AUC $\scriptstyle\uparrow$ & BS $\scriptstyle\downarrow$ & ECE $\scriptstyle\downarrow$ & AUC $\scriptstyle\uparrow$ & BS $\scriptstyle\downarrow$ & ECE $\scriptstyle\downarrow$ \\
\midrule
& \multicolumn{10}{c}{Olmo 3 7B, TriviaQA} \\
\cmidrule(lr){2-11}
GCM & \underline{0.446} & \underline{0.448} & \underline{0.331} & \underline{0.335} & 0.808 & 0.182 & 0.088 & 0.869 & 0.191 & 0.181 \\
\,\raisebox{0.5ex}{\(\llcorner\)} Post-hoc & \textbf{0.446} & \textbf{0.448} & \textbf{0.333} & \textbf{0.337} & \textbf{0.812} & \textbf{0.179} & 0.085 & 0.869 & 0.142 & \underline{0.035} \\
Oracle training & 0.379 & \underline{0.424} & 0.255 & 0.285 & \underline{0.795} & \underline{0.185} & \textbf{0.061} & \textbf{0.890} & \textbf{0.131} & \textbf{0.032} \\
\midrule
& \multicolumn{10}{c}{Olmo 3 7B, Jeopardy} \\
\cmidrule(lr){2-11}
GCM & \textbf{0.512} & \underline{0.523} & \underline{0.376} & \underline{0.382} & \underline{0.849} & \underline{0.159} & 0.079 & 0.828 & 0.152 & 0.120 \\
\,\raisebox{0.5ex}{\(\llcorner\)} Post-hoc & \underline{0.511} & \underline{0.522} & 0.347 & 0.352 & \underline{0.845} & \underline{0.161} & 0.085 & 0.828 & 0.135 & 0.038 \\
Oracle training & \underline{0.506} & \textbf{0.536} & \textbf{0.379} & \textbf{0.402} & \textbf{0.855} & \textbf{0.156} & \textbf{0.057} & \textbf{0.888} & \textbf{0.111} & \textbf{0.020} \\
\midrule
& \multicolumn{10}{c}{Marin 8B, TriviaQA} \\
\cmidrule(lr){2-11}
GCM & \underline{0.282} & 0.294 & 0.195 & 0.205 & 0.704 & 0.232 & 0.219 & 0.833 & 0.157 & 0.103 \\
\,\raisebox{0.5ex}{\(\llcorner\)} Post-hoc & \underline{0.287} & 0.294 & 0.180 & 0.189 & 0.705 & 0.225 & 0.224 & 0.833 & 0.143 & 0.046 \\
Oracle training & \textbf{0.349} & \textbf{0.505} & \textbf{0.252} & \textbf{0.373} & \textbf{0.840} & \textbf{0.163} & \textbf{0.105} & \textbf{0.901} & \textbf{0.110} & \textbf{0.028} \\
\midrule
& \multicolumn{10}{c}{Marin 8B, Jeopardy} \\
\cmidrule(lr){2-11}
GCM & 0.349 & 0.325 & 0.242 & 0.235 & 0.734 & 0.216 & 0.185 & 0.793 & 0.113 & 0.056 \\
\,\raisebox{0.5ex}{\(\llcorner\)} Post-hoc & 0.347 & 0.325 & 0.210 & 0.204 & 0.734 & 0.210 & 0.193 & 0.793 & 0.111 & 0.034 \\
Oracle training & \textbf{0.459} & \textbf{0.503} & \textbf{0.350} & \textbf{0.377} & \textbf{0.825} & \textbf{0.175} & \textbf{0.123} & \textbf{0.890} & \textbf{0.088} & \textbf{0.027} \\
\midrule
& \multicolumn{10}{c}{Olmo 3 32B, TriviaQA} \\
\cmidrule(lr){2-11}
GCM & 0.333 & 0.328 & 0.225 & 0.227 & 0.727 & 0.220 & 0.171 & 0.833 & 0.152 & 0.092 \\
\,\raisebox{0.5ex}{\(\llcorner\)} Post-hoc & 0.335 & 0.329 & 0.203 & 0.207 & 0.730 & 0.213 & 0.169 & 0.834 & 0.144 & 0.053 \\
Oracle training & \textbf{0.384} & \textbf{0.484} & \textbf{0.274} & \textbf{0.341} & \textbf{0.827} & \textbf{0.169} & \textbf{0.093} & \textbf{0.898} & \textbf{0.111} & \textbf{0.027} \\
\midrule
& \multicolumn{10}{c}{Olmo 3 32B, Jeopardy} \\
\cmidrule(lr){2-11}
GCM & 0.371 & 0.374 & 0.252 & 0.257 & 0.759 & 0.203 & 0.118 & 0.799 & 0.102 & 0.043 \\
\,\raisebox{0.5ex}{\(\llcorner\)} Post-hoc & 0.370 & 0.374 & 0.213 & 0.218 & 0.757 & 0.201 & 0.120 & 0.799 & 0.104 & 0.050 \\
Oracle training & \textbf{0.475} & \textbf{0.513} & \textbf{0.342} & \textbf{0.370} & \textbf{0.840} & \textbf{0.163} & \textbf{0.075} & \textbf{0.894} & \textbf{0.081} & \textbf{0.029} \\
\bottomrule
\end{tabular}
}
\end{center}
\end{table}

\begin{table}[t]
\caption{Transfer from TriviaQA or Jeopardy to BioASQ. The advantage of oracle training over other methods does not hold under domain transfer, consistent with prior work on the domain-specificity of correctness signals \citep{sky2024androids}.}
\label{tab:transfer-to-bioasq}
\begin{center}
\setlength{\tabcolsep}{8pt}
\small
\resizebox{\linewidth}{!}{%
\renewcommand{\arraystretch}{1.3}
\begin{tabular}{lcccccccccc}
\toprule
& \multicolumn{7}{c}{Contrast} & \multicolumn{3}{c}{Full} \\
\cmidrule(lr){2-8}\cmidrule(lr){9-11}
Method & $\Delta_0^b$ $\scriptstyle\uparrow$ & $\Delta_0$ $\scriptstyle\uparrow$ & $\Delta^b$ $\scriptstyle\uparrow$ & $\Delta$ $\scriptstyle\uparrow$ & AUC $\scriptstyle\uparrow$ & BS $\scriptstyle\downarrow$ & ECE $\scriptstyle\downarrow$ & AUC $\scriptstyle\uparrow$ & BS $\scriptstyle\downarrow$ & ECE $\scriptstyle\downarrow$ \\
\midrule
& \multicolumn{10}{c}{Olmo 3 7B, TriviaQA $\to$ BioASQ} \\
\cmidrule(lr){2-11}
Self-consistency & 0.122 & 0.137 & 0.075 & 0.083 & 0.602 & 0.260 & 0.121 & 0.738 & 0.216 & 0.087 \\
Non-oracle training & \textbf{0.223} & \underline{0.239} & \textbf{0.149} & \textbf{0.163} & 0.680 & 0.234 & 0.106 & \underline{0.805} & 0.189 & 0.090 \\
Oracle training & \underline{0.211} & \textbf{0.256} & 0.131 & \underline{0.155} & \textbf{0.691} & \textbf{0.223} & \textbf{0.072} & \textbf{0.805} & \textbf{0.184} & \textbf{0.054} \\
\midrule
& \multicolumn{10}{c}{Olmo 3 7B, Jeopardy $\to$ BioASQ} \\
\cmidrule(lr){2-11}
Self-consistency & 0.122 & 0.137 & 0.075 & 0.083 & 0.602 & \underline{0.260} & \textbf{0.121} & 0.738 & 0.216 & 0.087 \\
Non-oracle training & \underline{0.165} & 0.160 & \textbf{0.134} & \underline{0.132} & \underline{0.637} & 0.259 & 0.147 & \underline{0.778} & 0.201 & 0.090 \\
Oracle training & \textbf{0.183} & \textbf{0.198} & 0.123 & \textbf{0.132} & \textbf{0.645} & \textbf{0.249} & \underline{0.123} & \textbf{0.780} & \textbf{0.196} & \textbf{0.064} \\
\midrule
& \multicolumn{10}{c}{Marin 8B, TriviaQA $\to$ BioASQ} \\
\cmidrule(lr){2-11}
Self-consistency & \textbf{0.204} & 0.267 & \textbf{0.144} & \textbf{0.174} & 0.691 & 0.231 & 0.209 & 0.736 & 0.215 & \textbf{0.078} \\
Non-oracle training & 0.102 & \textbf{0.449} & 0.042 & \underline{0.153} & \textbf{0.781} & \textbf{0.199} & \textbf{0.163} & 0.739 & 0.253 & 0.187 \\
Oracle training & \underline{0.198} & 0.256 & \underline{0.129} & \underline{0.161} & 0.687 & 0.229 & 0.211 & \textbf{0.801} & \textbf{0.189} & \underline{0.081} \\
\midrule
& \multicolumn{10}{c}{Marin 8B, Jeopardy $\to$ BioASQ} \\
\cmidrule(lr){2-11}
Self-consistency & \textbf{0.204} & \underline{0.267} & \textbf{0.144} & \textbf{0.174} & \underline{0.691} & \underline{0.231} & \textbf{0.209} & 0.736 & 0.215 & \textbf{0.078} \\
Non-oracle training & \underline{0.151} & \textbf{0.334} & 0.032 & 0.087 & \textbf{0.694} & \textbf{0.225} & \underline{0.218} & 0.734 & 0.241 & 0.150 \\
Oracle training & \underline{0.144} & 0.028 & \underline{0.104} & 0.036 & 0.539 & 0.300 & 0.308 & \textbf{0.767} & \textbf{0.204} & 0.095 \\
\midrule
& \multicolumn{10}{c}{Olmo 3 32B, TriviaQA $\to$ BioASQ} \\
\cmidrule(lr){2-11}
Self-consistency & 0.139 & 0.151 & \underline{0.088} & \underline{0.099} & 0.616 & 0.258 & 0.151 & 0.751 & 0.201 & \textbf{0.048} \\
Non-oracle training & \underline{0.169} & 0.167 & 0.076 & 0.075 & 0.619 & \underline{0.246} & 0.152 & 0.774 & 0.206 & 0.108 \\
Oracle training & \textbf{0.198} & \textbf{0.207} & \textbf{0.112} & \textbf{0.121} & \textbf{0.651} & \textbf{0.241} & \textbf{0.135} & \textbf{0.805} & \textbf{0.180} & 0.055 \\
\midrule
& \multicolumn{10}{c}{Olmo 3 32B, Jeopardy $\to$ BioASQ} \\
\cmidrule(lr){2-11}
Self-consistency & \underline{0.139} & \underline{0.151} & \textbf{0.088} & \textbf{0.099} & \textbf{0.616} & \underline{0.258} & \textbf{0.151} & 0.751 & \underline{0.201} & \textbf{0.048} \\
Non-oracle training & \textbf{0.181} & \textbf{0.172} & \underline{0.081} & \underline{0.079} & \underline{0.615} & \textbf{0.250} & \underline{0.154} & 0.764 & 0.215 & 0.129 \\
Oracle training & \underline{0.169} & 0.134 & \underline{0.087} & 0.070 & \underline{0.591} & 0.265 & 0.176 & \textbf{0.770} & \textbf{0.198} & 0.077 \\
\bottomrule
\end{tabular}
}
\end{center}
\end{table}

\begin{table}[t]
\caption{Variance analysis (\cref{app:variance-analysis}).
Spearman correlations between checkpoint index and adapter variance for Olmo 3 7B on TriviaQA. For each question and layer, we compute the total variance (i.e.\ trace of the covariance matrix) of the model's hidden states across 10 confidence adapter training seeds, and we compute the correlation between this total variance and checkpoint index. We find that the mean/median correlation across questions is positive for all layers, suggesting that later checkpoints tend to have a larger space of viable confidence features that confidence adapters can pick up on.}
\label{tab:hidden-state-variance}
\centering
\small
\begin{tabular}{rcc}
\toprule
Layer & Mean $\rho$ & Median $\rho$ \\
\midrule
1 & 0.920 & 0.929 \\
2 & 0.914 & 0.929 \\
3 & 0.631 & 0.750 \\
4 & 0.507 & 0.536 \\
5 & 0.533 & 0.714 \\
6 & 0.535 & 0.750 \\
7 & 0.613 & 0.786 \\
8 & 0.532 & 0.607 \\
9 & 0.385 & 0.357 \\
10 & 0.414 & 0.429 \\
11 & 0.379 & 0.429 \\
12 & 0.409 & 0.429 \\
13 & 0.505 & 0.536 \\
14 & 0.467 & 0.464 \\
15 & 0.452 & 0.464 \\
16 & 0.442 & 0.429 \\
17 & 0.510 & 0.536 \\
18 & 0.488 & 0.536 \\
19 & 0.454 & 0.500 \\
20 & 0.397 & 0.464 \\
21 & 0.350 & 0.357 \\
22 & 0.312 & 0.321 \\
23 & 0.276 & 0.286 \\
24 & 0.279 & 0.286 \\
25 & 0.273 & 0.286 \\
26 & 0.274 & 0.286 \\
27 & 0.273 & 0.286 \\
28 & 0.278 & 0.286 \\
29 & 0.248 & 0.250 \\
30 & 0.178 & 0.143 \\
31 & 0.279 & 0.321 \\
32 & 0.674 & 0.679 \\
\bottomrule
\end{tabular}
\end{table}

\subsection{Empirical Support for Conditioning on Contrast Set Membership}
\label{app:learn-condition-conset}

To provide empirical support for our method of conditioning confidence estimates on contrast set membership (\cref{eq:condition-conset-independent}), we show that a fitted normalization function and \cref{eq:condition-conset-independent} produce similar results.

Suppose checkpoints 1 and 2 produce confidences $p_1$ and $p_2$ in their respective answers to a question that exactly one of them answers correctly.
We use the Bradley-Terry (BT) model \citep{bradley1952rank} to parameterize the space of functions that map $(p_1, p_2)$ to the probability that checkpoint 1 is correct.
The BT model can express a broad class of pairwise preferences functions \citep{sun2024rethinking} and has been widely used for preference optimization objectives \citep{christiano2017deep, ouyang2022training, rafailov2023direct}.

For a given confidence estimator and a given contrast set, we fit a model to predict $r(p_1, p_2)$, the probability that checkpoint 1 is correct for a given contrast-set question.
Importantly, we fit the model using the evaluation data (confidences and correctness labels), so the downstream results should be considered generous.
The BT model parameterizes $r$ with a utility function $w(p) > 0$, where we can ensure positivity with a reparameterized $h(p) = \log w(p)$:
\begin{align*}
    r(p_1, p_2) = \frac{w(p_1)}{w(p_1) + w(p_2)} = \sigma(h(p_1) - h(p_2))
\end{align*}
For example, \cref{eq:condition-conset-independent} uses $h(p) = \operatorname{logit}(p) = \log(\frac{p}{1-p})$.
Since $r$ is invariant to additive constants on $h$, we can set $h(0.5) = 0$.
Furthermore, we constrain $h$ to be antisymmetric about 0.5, i.e.\ $h(1-p) = -h(p)$, since negating both claims (i.e.\ the checkpoints' claims that their answers are correct) should give $r(1 - p_1, 1 - p_2) = 1 - r(p_1, p_2)$.
Thus, we fit $h(p)$ for $p \ge 0.5$, and define $h(p)$ for $p < 0.5$ by $h(p) = -h(1-p)$.
We fit a spline with knots at $0.5 = c_0 < c_1 < \ldots < c_K < 1$ using $K=30$ knots.
Each knot $c_k$ is chosen as the $k / (K+1)$ quantile of the empirical distribution of $\max(p, 1 - p)$ over the $2N$ confidence estimates produced in the given contrast set by the given confidence estimator, where $N$ is the number of questions in the contrast set.
We linearly interpolate between the knots, and define the slope between $c_K$ and 1 to equal that between $c_{K-1}$ and $c_K$.
We fit $h(c_0), \ldots, h(c_K)$ to minimize the negative log likelihood of $r$ -- a convex objective with respect to $h(c_0), \ldots, h(c_K)$ -- using LBFGS with 50 maximum iterations.

\cref{tab:oracle-vs-methods-unabridged-learn1d} shows the results using the fitted normalization functions.
We find the same trends from \cref{tab:oracle-vs-methods-unabridged}: confidence estimation methods underperform oracle training, showing that our findings are robust to the specific implementation of normalization.

\subsection{Oracle Training Versus Existing Methods}
\label{app:oracle-vs-methods-unabridged}

\cref{tab:oracle-vs-methods-unabridged} is an expanded version of \cref{tab:oracle-vs-methods-abridged} showing the same trends of headroom between existing confidence estimation methods and oracle training.

\cref{tab:methods-vs-ablations-unabridged} is an expanded version of \cref{tab:methods-vs-ablations-triviaqa} comparing SC and non-oracle training to their surrogate and copy ablations.
Based on contrast-set AUC, SC outperforms its ablations in 5/6 cases, while non-oracle training outperforms its ablations in only 3/6 cases.
We describe one intuition for why the surrogate ablations achieve higher contrast-set AUC than one might expect, given that they use an earlier checkpoint that often did not predict the correct answer.
Although different checkpoints are generally not independent with regard to their predictions, it may be instructive to approximate them as such, especially for questions where some of them made different predictions.
Since the checkpoints are similarly performant (\cref{tab:checkpoint-training-tokens}), ensembling their predictions by taking the majority vote can be expected to achieve a higher accuracy than any individual checkpoint \citep{huang2017snapshot, garipov2018loss}.
This would suggest that, for a question on which two evaluation checkpoints produce differing predictions, a third checkpoint (i.e.\ the surrogate checkpoint) can distinguish the correct answer from the incorrect answer with nontrivial probability, leading to a nontrivial contrast-set performance.
Intuitively, different checkpoints' incorrect predictions tend to differ, so even if the surrogate checkpoint predicted an incorrect answer, it is likely to consider the correct answer more plausible than an incorrect answer that is different from its own.

\subsection{Ablation of Multi-Checkpoint and Multi-Answer Training}
\label{app:ablation-multi-ckpt-multi-answer}

\cref{tab:mvs-ckpt-ans-triviaqa,tab:mvs-ckpt-ans-jeopardy} show that multi-checkpoint and multi-answer training often lead to substantial improvements in calibration.

\cref{tab:multi-ckpt-vs-ablations-olmo7b-jeopardy} echoes \cref{tab:multi-ckpt-vs-ablations-olmo7b-triviaqa} in showing that multi-checkpoint training outperforms its ablations, showing that its performance gain comes from both more data and access to multiple checkpoints' weights.
Furthermore, in \cref{tab:multi-ckpt-vs-ablations-olmo7b-jeopardy}, the single-checkpoint data ablation matches the unablated multi-checkpoint training on the class-balanced metrics $\Delta^b_0, \Delta^b$, indicating that most of the performance gain comes from viewing multiple checkpoints' weights to learn confidence features that generalize across checkpoints.

\subsection{GCM Results}
\label{app:gcm-results}

\cref{tab:oracle-vs-gcm-posthoc} shows all evaluation metrics for the GCM results summarized in \cref{fig:gcm}.
On full-set AUC, GCM consistently falls short of oracle training, especially for larger models.
On contrast-set AUC, GCM matches oracle training only on the smallest model (Olmo 3 7B).
We attribute GCM's nontrivial contrast-set AUC to the GCM knowing the correct answers to many contrast-set questions, enabling it to distinguish correct and incorrect answers without needing access to the internal states of the checkpoints that generated the answers.
This is supported by \cref{fig:gcm}, which shows that for the larger models (Marin 8B and Olmo 32B), the GCM backbone model (Qwen3-8B) has an increased error rate on contrast-set questions, and its gap from the oracle grows substantially.

\subsection{Domain Transfer}
\label{app:domain-transfer}

\cref{tab:transfer-to-bioasq} shows that oracle training and other methods drop to similarly low performances when transferring from TriviaQA or Jeopardy to BioASQ (biomedical domain).
This clarifies the scope of the oracle's advantage in our experiments as being limited to the training domain, consistent with prior work on the domain-specificity of correctness signals \citep{sky2024androids, srey2026signals, ying2026truthfulness}.

\subsection{Transfer Asymmetry Analysis}
\label{app:transfer-asymmetry}

\begin{figure}[t]
    \centering
    \includegraphics[width=0.9\linewidth]{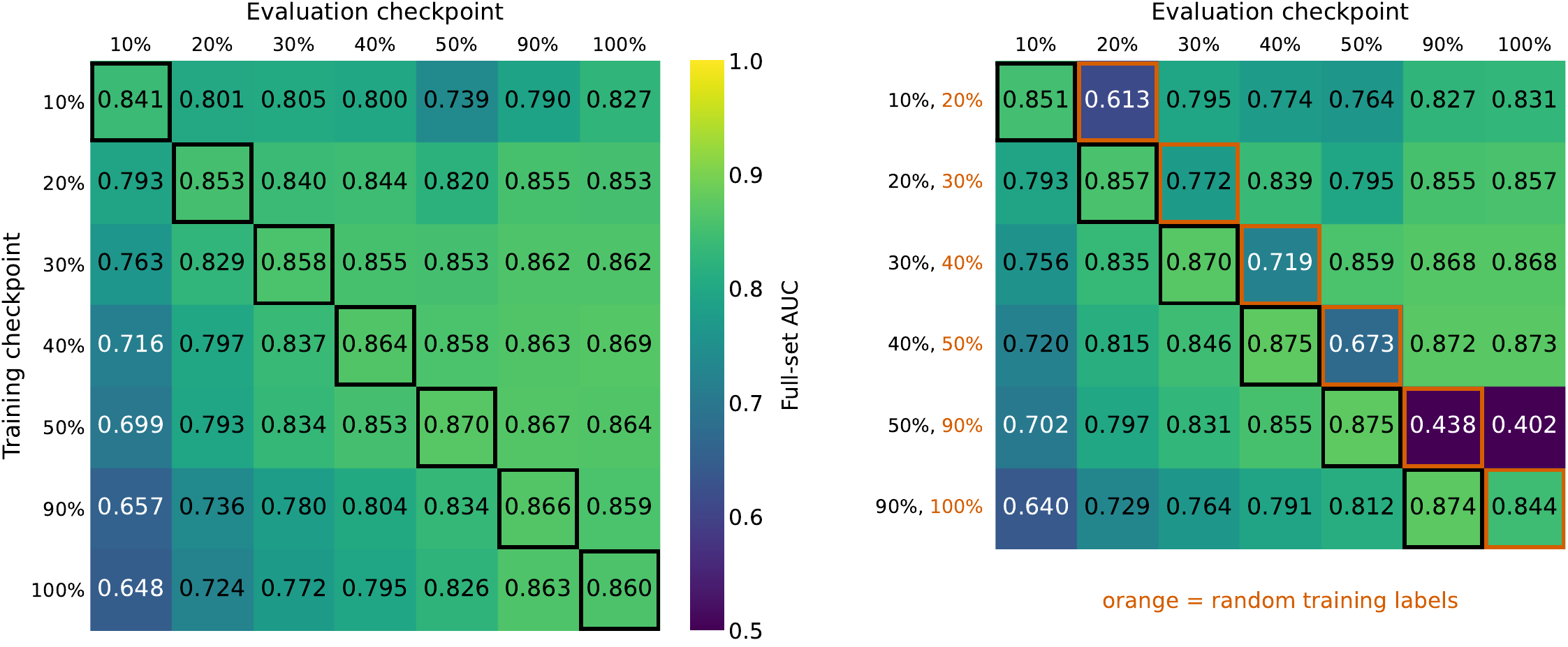}
    \caption{\textbf{(Left)} Transfer between pairs of training and eval checkpoints (Olmo 3 7B, Jeopardy, see \cref{fig:triviaqa-transfer} for TriviaQA). Forward transfer (top right) is strong, while backward transfer (bottom left) is weak.
    \textbf{(Right)} Transfer of multi-checkpoint training, where the second checkpoint gets random labels. The second-checkpoint AUC generally plummets, at no penalty to the first checkpoint.
    }
    \label{fig:jeopardy-transfer}
\end{figure}

\cref{fig:jeopardy-transfer} shows the same trends as \cref{fig:triviaqa-transfer}: Forward transfer is far more robust than backward transfer, and we can achieve catastrophic performance on a later checkpoint at no penalty to an earlier checkpoint.

\subsection{Cross-Seed Variance on Different Checkpoints}
\label{app:variance-analysis}

We further investigate the hypothesis in \cref{sec:analysis} that later checkpoints have developed a larger space of confidence features for confidence adapters to choose from.
Our hypothesis predicts that later checkpoints would have greater variance among separately trained confidence adapters.
To quantify this, for each checkpoint $k \in K$, we train 10 confidence adapters $f_{k,s}$ with different seeds $s \in S$.
We run each adapter $f_{k,s}$ on each evaluation question $q \in Q$ (using the respective checkpoint $k$ and checkpoint $k$'s predicted answer), storing the hidden state $h_{k,s,q,l}$ at each layer $l \in L$.
For each question $q$ and layer $l$, we compute the Spearman correlation $\rho_{q,l}$ between checkpoint index $k$ and total variance (i.e.\ trace of the covariance matrix) of the hidden states $\{h_{k,s,q,l}\}_{s \in S}$.
Importantly, the total variance is computed within a given question $q$, so the heterogeneity of questions in the dataset is not represented.
The correlation tending to be high would mean that later checkpoints tend to have a higher total variance.
For each layer $l$, we compute the mean and median of the Spearman correlations $\{\rho_{q,l}\}_{q \in Q}$.
\cref{tab:hidden-state-variance} shows that the mean correlation and the median correlation are positive for all layers, suggesting that later checkpoints tend to have a larger space of viable confidence features that confidence adapters can pick up on.

\section{Related Work}
\label{app:related-work}

\myparagraph{Contrast Sets.} Since neural networks are prone to learning spurious features for underspecified tasks \citep{geirhos2020shortcut, d2022underspecification}, a line of work has investigated training or evaluating them on minimal pairs -- instances with slight differences that affect the correct prediction -- to expose reliance on spurious features \citep{shekhar2017foil, marvin2018targeted, rudinger2018gender, sakaguchi2021winogrande}.
\citet{gardner2020evaluating} define a \textit{contrast set} to contain minimal pairs and show that models' performance systematically drops on human-written contrast sets for several language tasks.
Subsequent work has leveraged task-specific structure to automate the creation of contrast sets \citep{li2020linguistically, bitton2021automatic, huang2025math, mirzadeh2025gsm}.
Recently, \citet{ashuach2026masked} use contrast sets derived from disagreements between models to evaluate privileged knowledge of LLM introspection, finding that self-probes outperform peer-probes in some domains.
In this paper, we induce minimal pairs in knowledge by continually training a model and evaluate whether a confidence estimator faithfully tracks evolving knowledge.

\myparagraph{Metacognition.} Endowing AI models with metacognition -- awareness of their own knowledge and behavior -- would make them more transparent and auditable, promoting safety and alignment \citep{yona2026position, steyvers2026metacognition}.
For example, they would be able to communicate uncertainty upon identifying gaps in their knowledge, as well as seek out information to fill those gaps, which is relevant for active and continual learning \citep{gureckis2012self}.
As a step towards generalizable metacognition, \citet{binder2025looking} and \citet{guo2026introspective} show that a model trained to explain its behavior will explain its \textit{new} behavior after we change the model to behave differently.
Our results contribute to this line of work by showing that a confidence estimator trained to reflect one checkpoint's knowledge can nontrivially reflect a future checkpoint's knowledge even when the two knowledge sets differ, although this is not consistent across models.
Since persistent calibration requires faithfully tracking knowledge, we hope that it serves future work as a testbed for metacognition.

\end{document}